\documentclass[10pt,journal,compsoc]{IEEEtran}

\usepackage{graphicx}
\usepackage{amsfonts}
\usepackage{mathrsfs}
\usepackage{amssymb}
\usepackage{amsmath}
\usepackage{blkarray}
\usepackage[dvipsnames,table]{xcolor}
\newcolumntype{L}{>{\raggedright\arraybackslash}X}
\newcolumntype{C}[1]{>{\centering\let\newline\\\arraybackslash\hspace{0pt}}m{#1}}
\newcolumntype{S}{>{\hspace{-2ex}}}
\newcolumntype{Q}{>{\hspace{-2.5ex}}}
\usepackage{textcomp}
\usepackage{subfig}
\usepackage[utf8]{inputenc}
\usepackage{nccmath}
\usepackage{booktabs}
\newtheorem{definition}{Definition}

\usepackage{lineno,hyperref}
\modulolinenumbers[5]

\ifCLASSOPTIONcompsoc
  \usepackage[nocompress]{cite}
\else
  \usepackage{cite}
\fi
\ifCLASSINFOpdf
\else
\fi

\begin{document}

\title{Embedding Graph Convolutional Networks in Recurrent Neural Networks for Predictive Monitoring}

\author{Efrén~Rama-Maneiro,
        Juan~C.~Vidal,
        and~Manuel~Lama
        \IEEEcompsocitemizethanks{
          \IEEEcompsocthanksitem Efrén Rama-Maneiro, Juan C. Vidal, and Manuel Lama are with the Centro Singular de Investigación en Tecnoloxías Intelixentes (CiTIUS), Universidade de Santiago de Compostela, Santiago de Compostela, Spain. Juan C. Vidal is also with the Departamento de Electrónica e Computación, Universidade de Santiago de Compostela, Galicia, Spain. email: \{efren.rama.maneiro, juan.vidal, manuel.lama\}@usc.es \protect \\
        }
\thanks{Manuscript received \today~revised X}}

\markboth{IEEE Transactions on Knowledge and Data Engineering}%
{Rama-Maneiro \MakeLowercase{\textit{et al.}}: Embedding Graph Convolutional Networks in Recurrent Neural Networks for Predictive Monitoring}

\IEEEtitleabstractindextext{%
\begin{abstract}
Predictive monitoring of business processes is a subfield of process mining that aims to predict, among other things, the characteristics of the next event or the sequence of next events.  Although multiple approaches based on deep learning have been proposed, mainly recurrent neural networks and convolutional neural networks, none of them really exploit the structural information available in process models.  This paper proposes an approach based on graph convolutional networks and recurrent neural networks that uses information directly from the process model. An experimental evaluation on real-life event logs shows that our approach is more consistent and outperforms the current state-of-the-art approaches.
\end{abstract}

}

\maketitle
\IEEEdisplaynontitleabstractindextext
\IEEEpeerreviewmaketitle

\section{Introduction}
\begin{figure}
  \centering
  \includegraphics[width=0.45\textwidth]{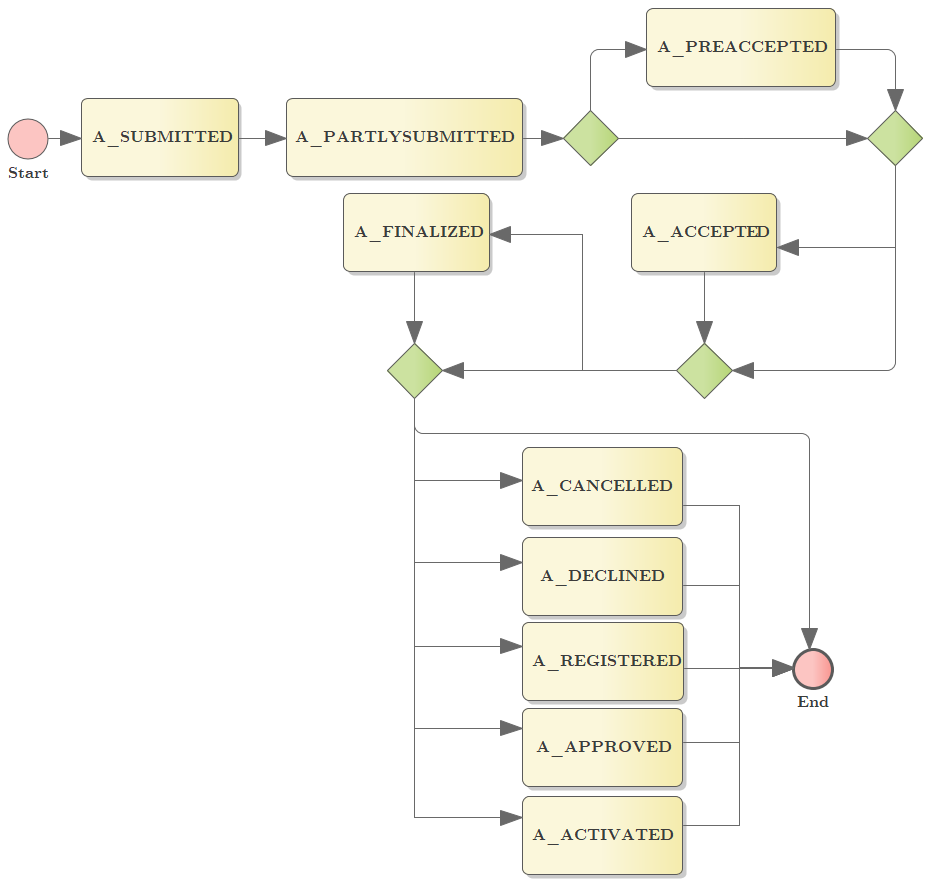}
  \caption{BPMN process model of the BPI 2012 A event log. Rectangles represent activities and the diamonds represent a decision between two activities.}\label{fig:bpmn-model}
\end{figure}

Process mining~\cite{Aalst2012} is a discipline that aims to analyze event logs by describing what has happened and what may happen in a business process given the information available in event logs. These event logs are records of the execution of a business process, which, in turn, can be defined as a series of activities performed by a set of resources to achieve a goal~\cite{Kirchmer2017}. The event logs are composed of events, which are identified by a case identifier, the activity performed, and a timestamp. Furthermore, they also can have case attributes, which are shared by the events of the same case, or event attributes, which are specific to each event. The sequence of events from the same case is called a trace, and if the sequence of events is still ongoing is called a prefix. There are multiple analysis that can be applied to event logs, such as discovering a process model (\textit{process discovery}), improve a process model using the information from the event log (\textit{process enhancement}), or compare a process model with an event log to check its degree of conformance (\textit{process conformance})~\cite{Aalst2012}. In particular, process discovery aims to ease the comprehension and analysis of business processes, a process model can be mined using a process discovery algorithm. This process model is an abstract representation of the underlying business process using the information of the event log. To illustrate these concepts let \tablename~\ref{tab:log} be an excerpt from an event log from the finance domain~\cite{Dongen2012}, namely the BPI 2012 A subprocess~\cite{Dongen2012}. Each row of the table is a different event, and, apart from the mandatory information of each event, the resource that executes each activity is also available. \figurename~\ref{fig:bpmn-model} shows the process model from the aforementioned event log in BPMN notation.

Predictive monitoring is a subfield of process mining concerned with forecasting how an ongoing case is going to unfold in the future~\cite{Maggi2014}.  These predictions may involve information such as what will be the next activity or set of activities, when the next event will happen, or how much time is left until the end of the case. Many machine learning techniques have been used to learn a predictive model of the aforementioned problems, although approaches based on deep learning are the ones that have obtained the best results~\cite{Tax2018}. Recurrent neural networks (RNNs), and, in particular, LSTMs (Long Short-Term Memory), are the most popular in this domain due to the sequential nature of traces in processes~\cite{RamaManeiro2020}.  However, autoencoders~\cite{Mehdiyev2017,Mehdiyev2018}, generative adversarial networks (GANs)~\cite{Taymouri2020}, convolutional neural networks (CNNs)~\cite{Mauro2019,Pasquadibisceglie2019}, or other types of RNNs~\cite{Hinkka2019,Khan2018,Heinrich2021} have also been used.

Almost all deep learning approaches in predictive monitoring build a predictive model exclusively from the information available in the traces of the processes~\cite{Evermann2017,Tax2017,Khan2018,Camargo2019,Hinkka2019,Pasquadibisceglie2019,Mauro2019,Zararah2021,Nguyen2019,Mehdiyev2017,Mehdiyev2018,Taymouri2020}. These approaches disregard the explicit structural information available in the process models, i.e, behavioral patterns such as loops and parallels may be missing since the neural networks might not be able to detect them with only the trace information. On the other hand, the approaches that rely on process models as input~\cite{Theis2019,Venugopal2021,Weinzierl2021} obtain information from the connectivity between the model activities (\cite{Venugopal2021,Weinzierl2021}) or by performing a token replay over its model (\cite{Theis2019}). However, they do not fully take advantage of the dependencies between the events because the whole prefix is encoded into a single vector where features of each execution state of the process can be overwritten when a loop occurs.

In this paper, we hypothesize that the performance of predictive models can be improved by using the information of both the traces and the process model, including explicitly information about each execution state of the process. Thus, the process model should facilitate the detection of some behavioral patterns common in process models, mainly loops and parallels. As far as loops are concerned, the neural network might benefit of knowing beforehand where the loop is going back, which activities compose the loop, and whether there exist inner loops or not. Regarding parallels, in which a set of activities might appear in any order in a trace, they are harder to detect without considering the process model, so knowing their presence beforehand could further ease the training phase of the neural network. Moreover, using the process model as an additional input could help the neural network to focus more on the information available in the event log since the relationships between the activities are already explicitly available from the process model.

\begin{table}
  \scriptsize
  \centering
\begin{tabular}{|lScSlSlSl|}
\hline
Trace ID & Event ID & Activity & Timestamp & Resource \\ \hline
214364 & $E_1$ & A\_SUBMITTED & 01/03/2012 & Joseph \\ \hline
214364 & $E_2$ & A\_PARTLYSUBMITTED & 02/03/2012 & Joseph \\ \hline
214364 & $E_3$ & A\_PREACCEPTED & 03/03/2012 & Joseph \\ \hline
214364 & $E_4$ & A\_ACCEPTED & 10/03/2012 & Enrico \\ \hline
214364 & $E_5$ & A\_FINALIZED & 11/03/2012 & Enrico \\ \hline
\end{tabular}
\caption{Excerpt of the BPI 2012 A event log (timestamps have been modified).}~\label{tab:log}
\end{table}

Taking this into account, we propose a predictive monitoring approach that leverages the information about both the process model and the trace by combining graph neural networks (GNNs) and RNNs. In our approach GNNs extract the structural information of the process model while RNNs deal with both the temporal information available in the traces and the evolution of the model execution state. Note that GNNs combine the graph representation of the process model with the features of its execution state for each event of the trace. Furthermore, our approach takes advantage of the full sequence of states and not only on the state related to the last event previous to the prediction. We have validated our approach using 10 publicly available datasets against 10 other state-of-the-art approaches. Results show that our approach improves the accuracy of predictions in almost all cases, confirming that using structural information of the process as an input of predictive models improves the convergence of the neural network, allowing to learn more effectively.

This paper is structured as follows, Section~\ref{sec:related-work} presents the state-of-the art in predictive monitoring techniques based on deep learning, Section~\ref{sec:preliminaries} shows some definitions and background needed to build the approach, Section~\ref{sec:approach} describes the proposed approach, Section~\ref{sec:evaluation} shows the evaluation of the proposed approach in real life event logs, and Section~\ref{sec:conclusions-future-work} highlights the conclusions of the paper and future work.

\section{Related work}
\label{sec:related-work}
Many predictive monitoring works are based on RNNs. Tax et al.~\cite{Tax2017} uses LSTM neural networks to predict both the next activity and the next timestamp in an ongoing process instance. The activities are encoded as a one-hot vector, and they also consider time features such as the time passed since the previous event or since the beginning of the case. Evermann et al.~\cite{Evermann2017} also aims to predict the next activity by encoding the activity through embeddings, instead of a one hot vector, thus reducing the dimensionality of the input. Khan et al.~\cite{Khan2018} uses a Differentiable Neural Computer~\cite{Graves2016}, which is a type of neural network that has an external memory to enhance the representation of longer term dependencies in a sequence. They define an encoder-decoder that is trained to predict either the next activity or the next timestamp. Jalayer et al.~\cite{Jalayer2020} also uses an encoder-decoder network, but they rely on an attention mechanism to take into account every hidden state of the LSTM that learns from the input sequence. Other predictive monitoring approaches propose novel ways to encode the attributes of the event log. Camargo et al.~\cite{Camargo2019} cluster the resources available in the event log into roles and learns a LSTM that simultaneously predicts the next activity, next timestamp, and next role. Hinkka et al.~\cite{Hinkka2019} also performs a clustering of the events based on their  attributes and use these clustering labels as additional information in a recurrent neural network.

There are also approaches that are based on convolutional neural networks. Pasquadibisceglie et al.~\cite{Pasquadibisceglie2019} uses CNNs to predict the next activity. They rearrange the prefixes in a grid-like fashion using an encoding in which for each event they count the number of occurrences present in the prefix. These prefixes are fed to a two-dimensional CNN which is used to predict the next activity. Mauro et al.~\cite{Mauro2019} also employs CNNs by adapting the Inception model~\cite{Szegedy2015} to predict the next activity, outperforming LSTMs in some event logs. In~\cite{Heinrich2021} Gated Convolutional Neural Networks~\cite{Dauphin2017} and Key-Value-Predict~\cite{Daniluk2017} attention networks are introduced. The former combines convolutional networks with a gating mechanism, which is similar as the one used in LSTMs and GRUs, while the latter tries to learn the correlation between pairs of elements that belong to the input sequence by means of an attention mechanism.

Furthermore, some approaches are testing novel architectures for predictive monitoring. Taymouri et al.~\cite{Taymouri2020} relies on GANs~\cite{Goodfellow2014} to predict the next activity and timestamp of an ongoing process instance, hypothesizing that the usage of this type of neural network would alleviate the need of high amounts of training data. Zaharah et al. \cite{Zararah2021} are focused on implementing Transformers~\cite{Vaswani2017} for predictive monitoring, which are a type of neural network that rely exclusively on attention mechanisms. This type of neural network avoid the predictive performance degradation when the RNN faces longs sequences and improves the learning phase training and inference speed.

Finally, some works rely on explicit process models to help encode the prefixes. Theis et al.~\cite{Theis2019} builds feature vectors by performing a token replay of a prefix over a process model. These feature vectors include the information about the most recent activation on a Petri net, the time of activation, which is a decay function, the number of tokens and the attributes available in the event log. These vectors are then fed to a deep feed-forward neural network. In~\cite{Weinzierl2021}, the usage of Gated Graph Neural Networks is explored. Their adjacency matrix can be based on the events of a prefix, on the activities of the prefix or on the Directly Follows Graph (DFG) extracted from the event log. The edges in their graph convey information about the prefix. Venugopal et al.~\cite{Venugopal2021} investigate the usage of GCNs for predicting the next activity. Their adjacency matrix is based on the DFG, and they focus on the differences of various methods of encoding this adjacency matrix (binary, weighted, with a Laplacian transform, etc.).

However, none of the aforementioned approaches really exploit the structural information available in the process model. Even though \cite{Theis2019} relies on Petri nets, they do not fully leverage the interactions between the activities available in the process model, since they focus exclusively on the last state after the token-replay of the prefix over the model, disregarding the interactions between the events of the prefix. On the other hand, \cite{Venugopal2021} and \cite{Weinzierl2021} rely on Directly Follows Graphs (DFGs), which are much less expressive than Petri nets. Furthermore, they are still subjected to overwriting information in their encoding when a loop occurs in the prefix, since they rely on the last state of the model when building their feature vectors. Instead, our approach takes advantage of the information available in the Petri net representation of the process model, by using both the information available in the full sequence of events of and its corresponding states of execution of the process model.

\section{Preliminaries}
In this section, we present the main concepts needed to understand our approach for predicting the next activity of a running case.
\label{sec:preliminaries}
\subsection{Definitions}

\begin{definition}[event]
  \label{def:event}
  Let $A$ be the universe of activities, $C$ the universe of cases, $T$ the time
  domain, and $D_1, \ldots, D_m$ the universes of each of the attributes of the
  traces and events of the event log, with $m \geq 0$. An event $e \in E$ is a tuple
  $(a, c, t, d_1, \ldots, d_m)$ where $a \in A$, $c \in C$, $t \in T$ and $d_i
  \in \{D_i \cup \epsilon\}$ with $i \in \left[1, m\right]$ and $\epsilon$ being
  the empty element.
\end{definition}

Given definition~\ref{def:event}, we distinguish between event-level
attributes, which are specific of a given event, and trace-level attributes,
which are common for every event of the trace. In this paper, for simplicity, we
only deal with event-level attributes.

\begin{definition}[trace]
  Let $\pi_A$, $\pi_C$, $\pi_T$ and $\pi_{D_i}$ be functions that map an event to an activity, a case identifier, a timestamp, and an attribute, that is, $\pi_A(e) = a$, $\pi_C(e) = c$, $\pi_T(e) = t$, and $\pi_{D_i}(e) = d_i$. Also let $S$ be the universe of traces, then a trace $\sigma \in S$ is a non-empty sequence of events $\sigma = \langle e_1, \ldots, e_n \rangle$ which holds that $\forall e_i, e_j \in \sigma; i, j \in \left[1, n\right]: j > i \land \pi_C(e_i) = \pi_C(e_j) \land \pi_T(e_j) \geq \pi_T(e_i)$ where $|\sigma| = n$.
\end{definition}

\begin{definition}[event log]
An event log is a set of traces, $L = \{\sigma_1, \ldots, \sigma_l\}$ such as $L = \{\sigma_i | \sigma_i \in S \land i \in \left[1, l\right]\}$ where $|L| = l$.
\end{definition}

\begin{definition}[prefix]
Let $\sigma$ be a trace such as $\sigma = \langle e_1, \ldots, e_{n} \rangle$ and $k \in \left[1, n\right]$ be any positive integer. The event prefix of length $k$, $hd^k$ can be defined as follows: $hd^k(\sigma) = \langle e_1, \ldots, e_k \rangle $. The activity prefix can be defined as the application of $\pi_A$ to the whole event prefix, being $\pi_A(hd^k(\sigma)) = \langle \pi_A(e_{1}), \ldots, \pi_A(e_{k}) \rangle $ respectively.
\end{definition}

Recalling the example event log from \tablename~\ref{tab:log}, the event prefixes are: $E_1$, $E_1 \rightarrow E_2$, $E_1 \rightarrow E_2 \rightarrow E_3$, $E_1 \rightarrow E_2 \rightarrow E_3 \rightarrow E_4$, and $E_1 \rightarrow E_2 \rightarrow E_3 \rightarrow E_4 \rightarrow E_5$.

Predictive monitoring aims to forecast how a given running case will unfold in the future. In this paper, we will focus on predicting the next activity of a given prefix. Formally:

\begin{definition}[next activity prediction]
Let $hd^k(\sigma)$ be an event prefix such as $hd^k(\sigma) = \langle e_1,
\ldots, e_k \rangle$, $e'$ be a predicted event by a function $\Omega$ be the
concatenation operator between two sequences, then, the next activity prediction
problem can be defined as learning a function $\Omega_A$ such as
$\Omega_A(hd^k(\sigma)) = \pi_A(e'_{k+1})$.
\end{definition}

In this paper we leverage information from process models using deep learning. Our business processes models are represented by a place/transition Petri net~\cite{Desel1998}, because it allows a richer representation of the behavior of a business process than other approaches such as BPMN or DFG. A Petri net is a bipartite graph composed by two types of nodes: places, which can contain tokens that represent the state of the Petri net in a given moment, and transitions, which represent each of the activities of the process. More formally, a Petri net can be defined as follows:

\begin{definition}[Petri net]
A Petri net is a tuple $N = (P, T, F)$ where:
\begin{itemize}
  \item $P$ is a finite set of places.
  \item $T$ is a finite set of transitions.
  \item $P \cap T = \emptyset$
  \item $F \subseteq (P \times T) \cup (T \times P)$ is a set of directed arcs.
\end{itemize}
\end{definition}

\begin{figure}
  \centering
  \includegraphics[width=0.5\textwidth]{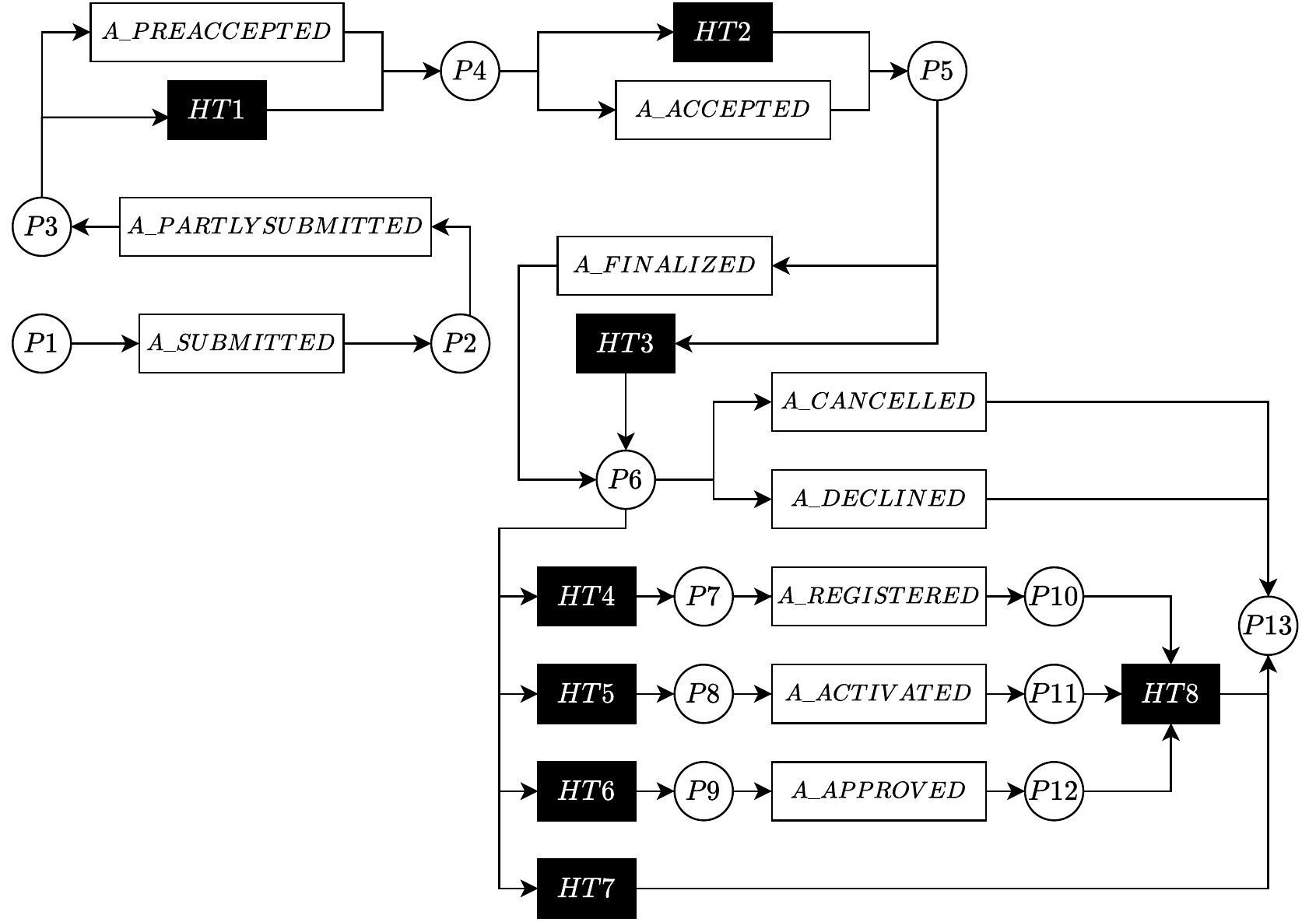}
  \caption{Process model mined as a Petri net from the BPI 2012 A event log. Circles represent the
  places, white rectangles are observable transitions, and black rectangles are
  silent transitions.}\label{fig:model}
\end{figure}

Note that transitions can be classified in two categories: \textit{observable transitions}, which are associated with a process activity and whose firing generates an event; and \textit{silent transitions}, which are related to control flow routing and whose firing does not correspond to the execution of an activity. \figurename~\ref{fig:model} shows the process model, represented as a Petri net, from the aforementioned BPI 2012 A event log from \tablename~\ref{tab:log}.

\begin{definition}[Marking]
The marking of a Petri net is a function $M : P \rightarrow \mathbb{N} $ that returns the number of tokens in a given place.
\end{definition}

\begin{definition}[Preset of a node of the Petri net]
Let $x \in T \cup P$ be a node of the Petri net graph, then the set of inputs of $x$, $\bullet x$, is defined by $\bullet x = \{y\ |\ (y,x) \in F \land ((x \in P \iff (y \in T \lor y = \epsilon)) \lor (x \in T \iff y \in P))\}$.
\end{definition}

\begin{definition}[Postset of a node of the Petri net]
Let $X \in T \cup P$ be a node of the Petri net graph, then set of outputs of $x$, $x \bullet$ is defined by $x \bullet = \{y\ |\ (x,y) \in F \land ((x \in P \iff (y \in T \lor y = \epsilon)) \lor (x \in T \iff y \in P))\}$.
\end{definition}

Petri nets of processes often have only one start place with no inputs ($\bullet s = \emptyset$) and an end place with no outputs ($e \bullet = \emptyset$). A transition can only be fired (i.e, is enabled) if all its input places contain a token. Formally:

\begin{definition}[Enabled transitions]
Let $t \in T$ be a transition. Then, we say that $t$ is enabled and, thus, can be fired if $\forall p \in \bullet t; M(p) > 0$.
\end{definition}

When a transition is fired, a token is deleted from each one of the input places of the transition and a token is created in each one of its output places. Formally:

\begin{definition}[Execution semantics]
Let $t \in T$ be an enabled transition. If $t$ is fired a new marking $M'$ is created such as $\forall p \in \bullet t, M'(p) = M(p) - 1$, and $\forall p \in t \bullet; M'(p) = M(p) + 1$.
\end{definition}

\begin{figure}
  \centering
  \includegraphics[width=0.5\textwidth]{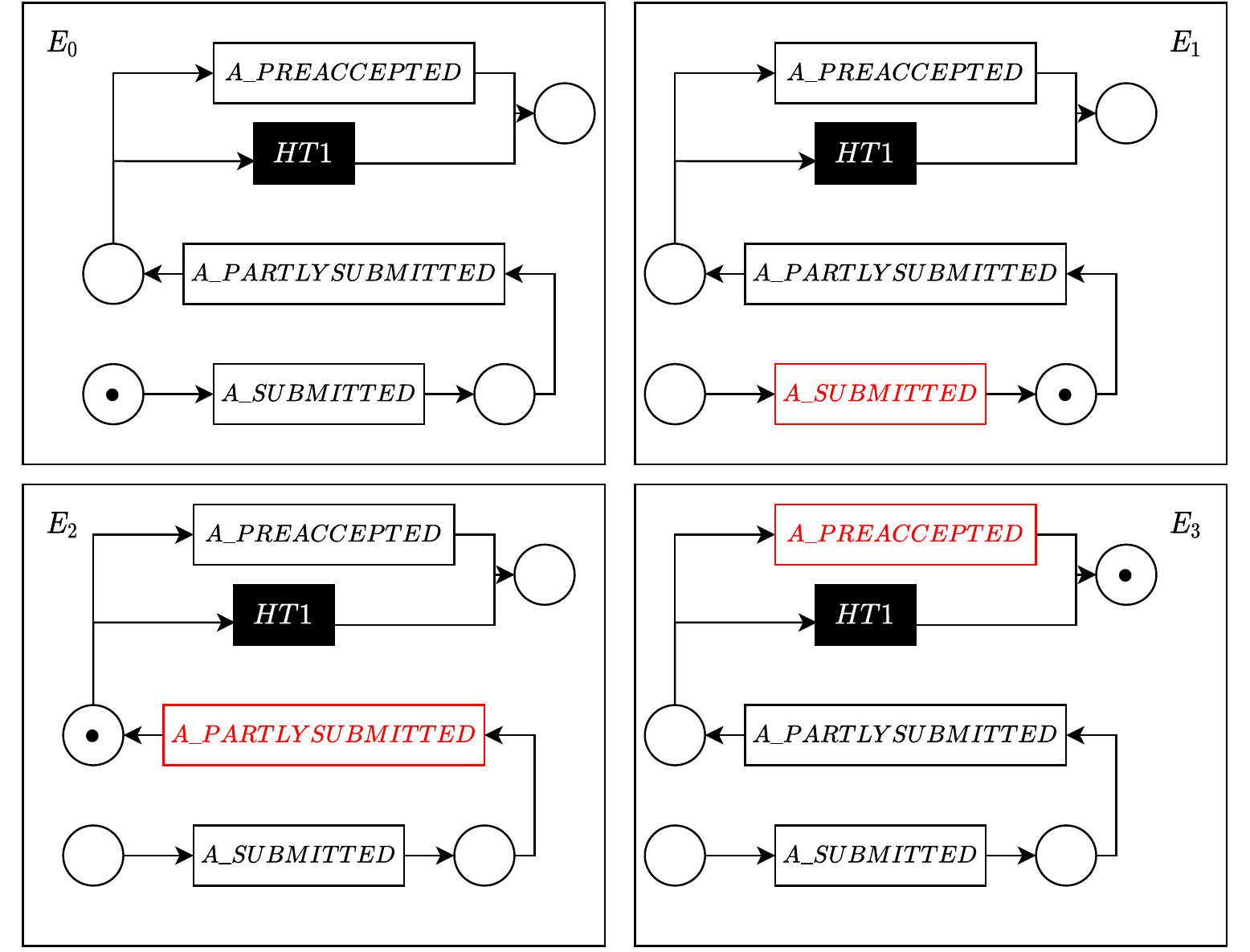}
  \caption{Replaying the first three events from the event log of \tablename~\ref{tab:log}. The event $E_0$ denotes the initial state of the Petri net.}\label{fig:execution}
\end{figure}

\figurename~\ref{fig:execution} shows how the first three events from the event log of \tablename~\ref{tab:log} are token-replayed (execution semantics) for part of the process model of \figurename~\ref{fig:model}. The transitions highlighted in red denote which transition is enabled in each moment. The initial marking before firing any transition of the Petri net is denoted as $E_0$.

Our model reduces the Petri net of the process model to a \textit{place graph} that is easier to handle in our deep learning-based solution.

\begin{definition}[Place graph]
  \label{def:place-graph}
  A place graph is a directed graph $G = (V, E)$ that represents a compact form of a Petri net $N = (P, T, F)$ where:
  \begin{itemize}
    \item $V$ is the set of vertices of the graph and is equivalent to the set of places of the Petri net $P$, i.e, $v \in V \iff p \in P$.
    \item $E$ is the set of edges of the graph. Each edge $e \in E$ connects two vertices $v \in V$ if and only if the two places that represent the vertices are interconnected by a transition in between, i.e., let $t \in T$; $p_1, p_2 \in P$; and, $v_1, v_2 \in V$; then $(v_1, v_2) \in E \rightarrow (p_1, t) \in F \land (t, p_2) \in F$.
  \end{itemize}
\end{definition}

\begin{figure}
  \centering
  \includegraphics[width=0.47\textwidth]{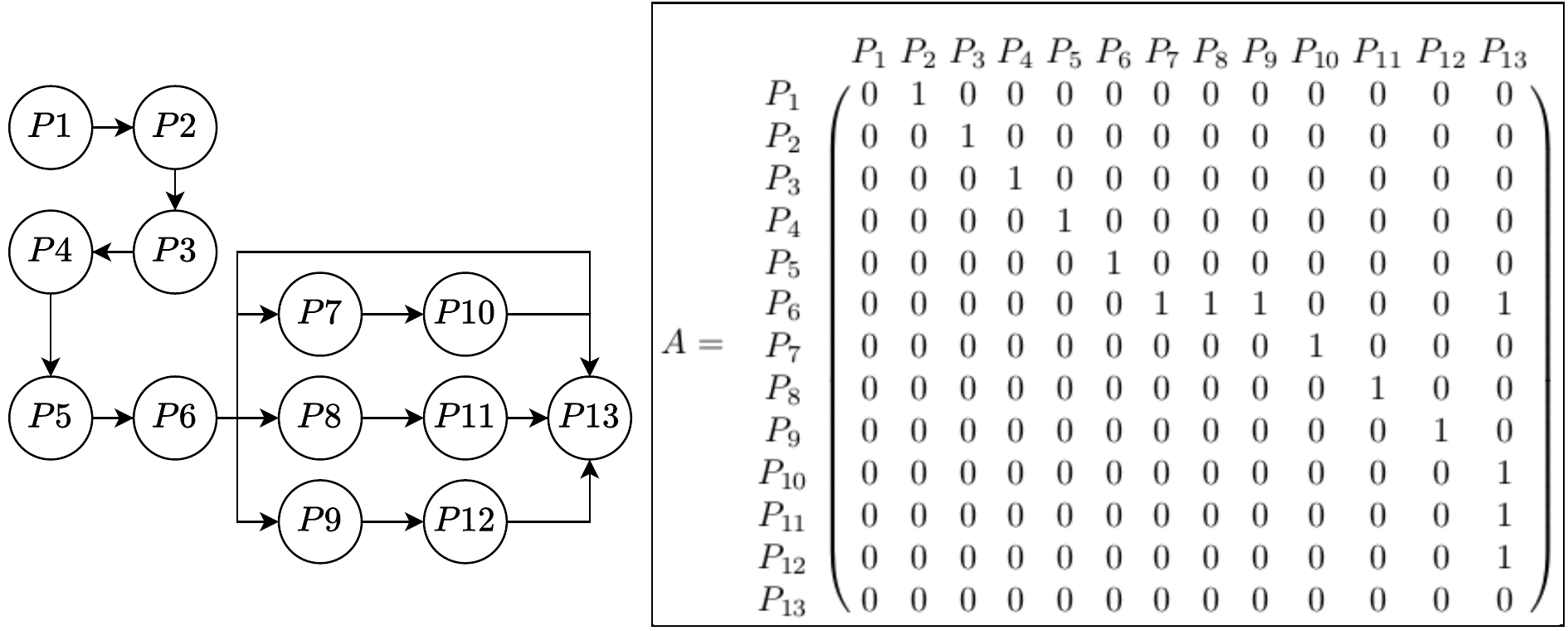}
  \caption{Place graph from the process model of \figurename~\ref{fig:model}. Adjacency matrix resulting from the process model.}~\label{fig:reduced-model}
\end{figure}

The place graph is represented in our approach by a binary adjacency matrix, i.e, $A \in \mathbb{R}^{|P| \times |P|}$, where $P$ are the places of the Petri net. The rationale for using a place graph instead of considering the full graph of the process model is twofold. First, the interactions between the activities of a Petri net can be represented only by the tokens consumed or produced on the places, because each event has associated a snapshot of the execution state of the Petri net. Second, the memory consumption is reduced by approximately a factor of two, which is important since the dimensions of every matrix related to the graph depend on the number of nodes of the graph. \figurename~\ref{fig:reduced-model} shows the place graph for the process model depicted in \figurename~\ref{fig:model}.

\subsection{Recurrent neural networks}
\label{sec:rnn}
Deep learning techniques attempt to model high-level data abstractions using multiple layers of neural networks. Of all available deep learning techniques, Recurrent Neural networks (RNNs)~\cite{Goodfellow2014} are one of the most popular architectures for predictive monitoring due to the sequential nature of the traces of an event log. This type of neural network contains cyclical connections in which the next state of the network depends on both the previous state of the network and the current input to the network. In their most basic form, this type of neural network suffers from vanishing gradients which makes them inappropriate to learning long sequences. Two of the most popular solutions to this problem are Long Short-Term Memory (LSTMs)\cite{Hochreiter1997} and Gated Recurrent Units (GRUs)~\cite{Cho2014}. These types of neural networks use a memory cell that has an internal recurrence as well as the usual recurrence of a regular RNN. In this paper, we rely on GRUs as the foundation block for our graph-recurrent architecture due to the easiness of computation and similar results obtained with respect to LSTMs. Formally, GRUs can be defined as follows:

\begin{equation}
  \begin{aligned}
  z_t &= \sigma(W_zx_t + U_zh_{t-1} + b_z) \\
  r_t &= \sigma(W_rx_t + U_rh_{t-1} + b_r) \\
  \tilde{h_t} &= tanh(W_h x_t + U_h(r_t \circ h_{t-1}) + b_z ) \\
  h_t &= z_t \circ h_{t-1} + (1 - z_t) \circ \tilde{h_t}
  \end{aligned}
\end{equation}

In the previous equations, $z$ refers to the ``update gate'', which controls the
amount of information that flows from the past to the future; $r$ is called the
``reset gate'', which filters how much information from the past is forgotten;
$\tilde{h}$ represents the calculation of the current memory; and $h_t$
corresponds to the final calculation of the memory of the cell, which can be
interpreted as how much information is retained from the past and how much
information is updated. 

The previously described model is the foundation block that we will use to build our graph recurrent predictive model.

\subsection{Graph neural networks}
\label{sec:gcn}

Unlike CNNs, which operate on a regular euclidean space, GNNs are a type of neural networks that operate on the graph structure, capturing dependencies between each of the nodes of a graph. Graph convolutional networks (GCN)~\cite{Kipf2017} are a type of GNN architecture that is defined as follows: 

\begin{equation}
  \label{eq:gcn-1}
  g_{\theta} \star x \approx \theta(I_N + D^{-\frac{1}{2}}AD^{-\frac{1}{2}}x)
\end{equation}

Where $I_N$ is the identity matrix, $A$ is the adjacency matrix of the graph, and $D$ is the degree matrix of the graph. Equation~\ref{eq:gcn-1} can lead to exploding/vanishing gradients, so the \textit{renormalization trick} $I_N + D^{-\frac{1}{2}}AD^{-\frac{1}{2}} \rightarrow \tilde{D}^{-\frac{1}{2}}\tilde{A}\tilde{D}^{-\frac{1}{2}}$ is introduced in~\cite{Kipf2017} by adding self-loops to the graph, i.e, $\tilde{A} = A + I_N$ and $\tilde{D}_{ii} = \sum_j \tilde{A}_{ij} $. Thus, we achieve the following GCN operator:

\begin{equation}
  \label{eq:gcn-2}
  GCN(A, X) = \tilde{D}^{-\frac{1}{2}}\tilde{A}\tilde{D}^{-\frac{1}{2}}X\Theta
\end{equation}

Where $\Theta$ is a matrix of learnable filter parameters. It should be noted that the expression~\ref{eq:gcn-2} does not take into account directionality in the input graph due to the symmetric normalization $\tilde{D}^{-\frac{1}{2}}\tilde{A}\tilde{D}^{-\frac{1}{2}}$, which is unsuitable for Petri nets, which are directed bipartite graphs. Thus, in this paper we modified the normalization to take into account directionality by using the random walk normalized Laplacian with the renormalization trick:

\begin{equation}
  \label{eq:gcn-3}
  GCN(A, X) = \tilde{D}^{-1} \tilde{A} X \Theta
\end{equation}

The output of a GCN operation is a matrix $\mathbb{R}^{N \times H}$, where $N$ is the number of nodes and $H$ is the hidden size. Since the objective of the approach is to predict the next activity of an ongoing process, we will assign a unique label to a whole graph (graph-level task). Thus, a summary operation over the graph is needed to project the output of a graph convolution operation into a single vector with a length equal to the number of activities. This operation is called \textit{readout}, and it reduces the graph outputs in the dimension of the nodes, i.e, the result of a readout operation is a vector $\mathbb{R}^{H}$. There are many readout operations available such as maximum, average, sum, weighted average using attention, etc.  In this paper we use the maximum operation for its simplicity and because it is suitable for classification problems~\cite{Xu2019}. Formally, let $h \in \mathbb{R}^{N \times H}$ be the matrix resulting from applying a graph convolution operation, then the maximum readout operation is defined as follows:

\begin{equation}
  h_G = max(h_{v})\ |\ v \in G
\end{equation}

Where $h_G$ is the summarized representation of the graph $G$, $max$ is the element-wise max-pooling operation, and $h_v$ is the learned embedding for the node $v$.

\subsection{Graph Recurrent Neural Networks}
\label{sec:grnn}
The GCN model presented in Section~\ref{sec:gcn} is unsuitable for predictive monitoring because it does not fully take into account the information available in the whole sequence of execution states related to the events. Thus, in this paper we propose to combine the equations from Section 3.1 and 3.2. We substitute the product between the learnable weights and the inputs --- or the hidden state --- by a graph convolution, leading to a Graph Recurrent Neural Network~\cite{Ruiz2020}. Thus, the equations for a GRU are rewritten as follows:

\begin{equation}
  \begin{aligned}
  z_t &= \sigma(GCN(x_t, A) + GCN(h_{t-1}, A) + b_z) \\
  r_t &= \sigma(GCN(x_t, A) + GCN(h_{t-1}, A) + b_r) \\
  \tilde{h_t} &= tanh(GCN(x_t, A) + r_t \circ GCN(h_{t-1}, A) + b_z ) \\
  h_t &= z_t \circ h_{t-1} + (1 - z_t) \circ \tilde{h_t}
  \end{aligned}
\end{equation} 

This allows to perform the graph convolution operation alongside the time dimension while performing the GRU operations. In this way, the neural network can take advantage of the encoding of the full sequence of state executions and mitigate the overwriting of the features when a loop occurs in the process, because the neural network can take advantage of all information available in the prefix.

\section{Approach}
\label{sec:approach}
\begin{figure*}
  \centering
  \includegraphics[width=0.8\textwidth]{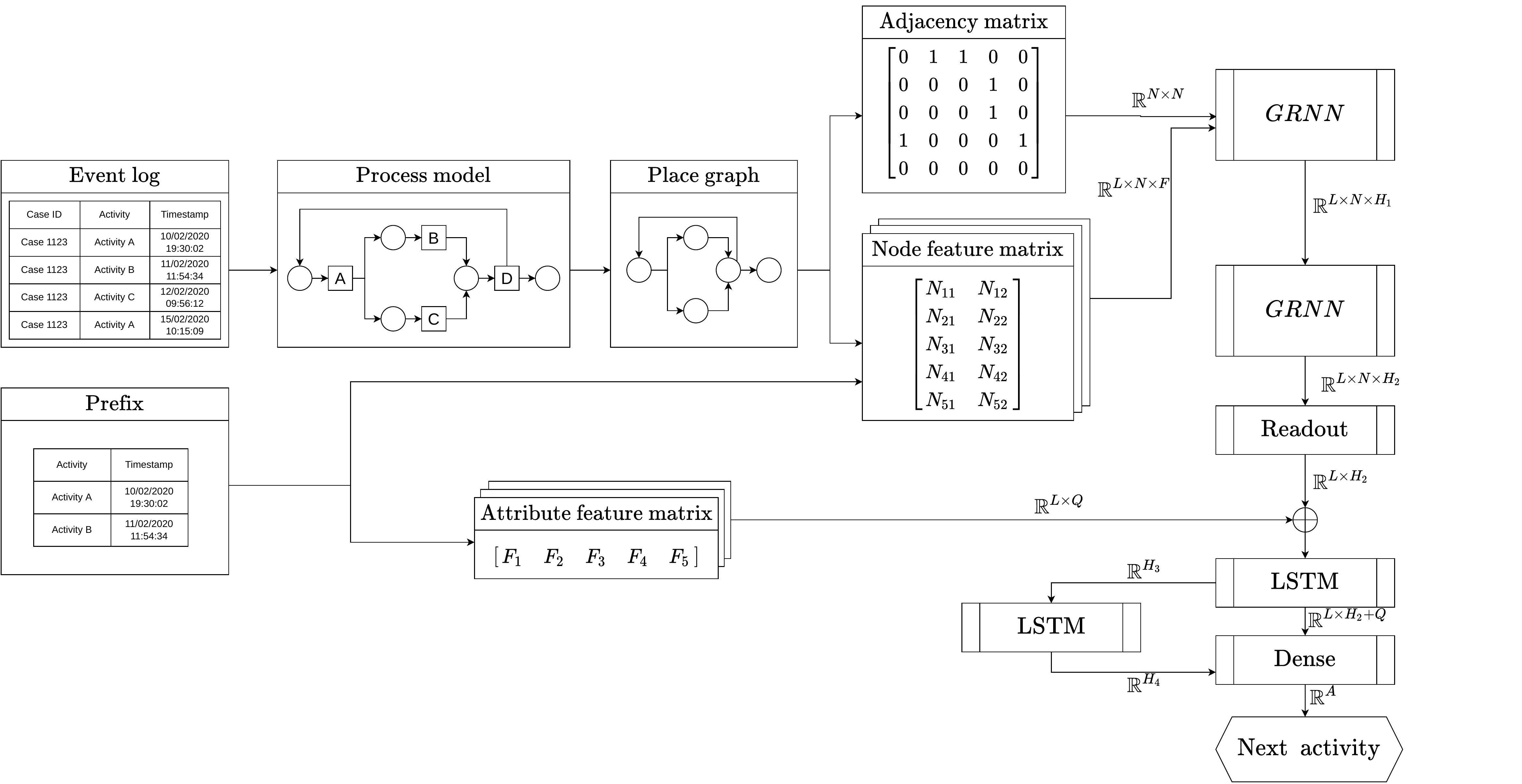}
  \caption{Pipeline of TACO. The sizes of the matrices are highlighted at each step, where $N$ is the number of places of the process model, $L$ is the length of the longest trace of the event log, $H_n$ is the hidden size of the layer $n$, $Q$ and $F$ are the dimensions of the features of the event log and the model, respectively, $A$ is the number of different activities of the event log, and $\oplus$ denotes the concatenation operator.}~\label{fig:pipeline}
\end{figure*}

In this section, we present our approach, ``Recurren\textbf{\underline{t}} Gr\textbf{\underline{a}}ph \textbf{\underline{C}}onvolutional Pr\textbf{\underline{o}}cess Predictor'' (\textbf{TACO}). TACO tries to solve a multiclass classification problem that could also be viewed as a supervised graph classification problem, mapping a label to a given graph structure and its replay. The steps to build the predictive model are depicted in \figurename~\ref{fig:pipeline}. First, a process model, represented by a Petri net, is mined from an event log using a discovery algorithm. We use the Split Miner~\cite{Augusto2018} discovery algorithm, even though our approach works with any discovery algorithm. We select the best model in terms of fitness by tuning two hyperparameters available in Split Miner, following the same approach as in~\cite{Theis2019}: the frequency threshold $\epsilon$, which controls the filtering process, and $\eta$, which controls the parallelism detection. Note that we do not require the process model to have a perfect fitness, i.e, that every trace can be replayed in the process model. If a prefix can not be replayed in the process model because of a misalignment of the trace, we fill the places with the tokens needed to fire the corresponding transition. 

After the process model is mined, we convert it into a place graph (see Definition~\ref{def:place-graph}), which will be represented as an adjacency matrix. 
The execution state of a place graph extends the Petri net marking to also include the features associated to each place. The matrix representation of this execution state will be called \textit{node feature matrix} (see Definition ~\ref{def:node-matrix}). 
Taking this into account, we can encode the structural information of the graph after the execution of each event of a prefix as a sequence of \textit{node feature matrices}. 
Furthermore, our approach also uses a sequence of matrices, called \textit{attribute feature matrix} (Definition ~\ref{def:attribute-matrix}), with the information contained in the events of the event log, such as activities, timestamps or attributes. All these feature matrices are further explained in Section~\ref{sec:encoding}.

TACO combines the sequence of node feature matrices with the adjacency matrix as input to a stacked GRNN. Then, an independent stacked LSTM has as input both the results of the GRNN and the attribute feature matrix. The results from the GRNN are summarized by means of a readout operation, which is the maximum of each characteristic learned by the GRNN for each node of the matrix. This operation reduces the results from a matrix of dimension $\mathbb{R}^{L \times N \times H_2}$ to a matrix $\mathbb{R}^{L \times H_2}$, where $H_2$ is the hidden size of the second layer of the GRNN. Finally, the readout results are fed to a softmax classifier to obtain the prediction of the next activity.

\subsection{Encoding}
\label{sec:encoding}
In this paper we distinguish between features that are specific for each node of the place graph and features that are inherent to each event of the prefix. The former features belong to a node feature matrix whereas the latter belong to a attribute feature matrix. Formally, a node feature matrix can be defined as follows:

\begin{definition}[Node feature matrix]
\label{def:node-matrix}
Let us define the marking of a place graph as a function $GM : P \rightarrow \mathbb{R}^{|Q|}$ that returns the cross product of features of a place, where $P$ is the set of places of the Petri net and $Q$ is the set of features. 
A node feature matrix is a matrix $A \in \mathbb{R}^{|P| \times |Q|}$ such that $(a_{ij}) = GM(p_i)_j$, where $p_i \in P$, $GM(p_i) = \{(r_1, \dots, r_{|Q|}): r_k \in \mathbb{R}, 1 \le k \le |Q|\}$, $1 \le i \le |P|$, and $1 \le j \le |Q|$. 
\end{definition}

The node feature matrix captures the marking of a place graph at a given time.  Let $A_n$ be the a state of node feature matrix of a place graph $P$, and let $R: GM \rightarrow GM$ be a function that changes a marking according to the execution semantics of Petri nets, e.g., using a token-replay function. Thus, the execution state of $P$ after replaying the next event can be defined as $A_{n+1} = R(A_n)$. Consequently, the node feature matrix that represents the token-replay of a prefix of size $n$ can be defined recursively from its previous states. Note that this formulation of the node feature matrix retains the information from previous events, due to the recursive definition of this encoding.


As aforementioned, each row of the node feature matrix captures the features associated to a place $p_i$. 
However, the number of columns/features depends on the structure of the Petri net.
In fact, we have as many columns as the number of transitions that can be fired concurrently in the Petri net, also accounting for every possible firing of hidden transitions.
Thus, this number can be defined as $max(|\bullet p_1|, ..., |\bullet p_n|) \; \forall \; n \in [1, |P|]$, where $p_n \in P$. 
Moreover, an additional column stores the place identifier when the place has a token at the current marking. 
Although this information is redundant, as this identifier represents the row place $p_i$, it improves the results of the GRNN. 


\begin{figure}[htb]
  \centering
    \includegraphics[width=0.49\textwidth]{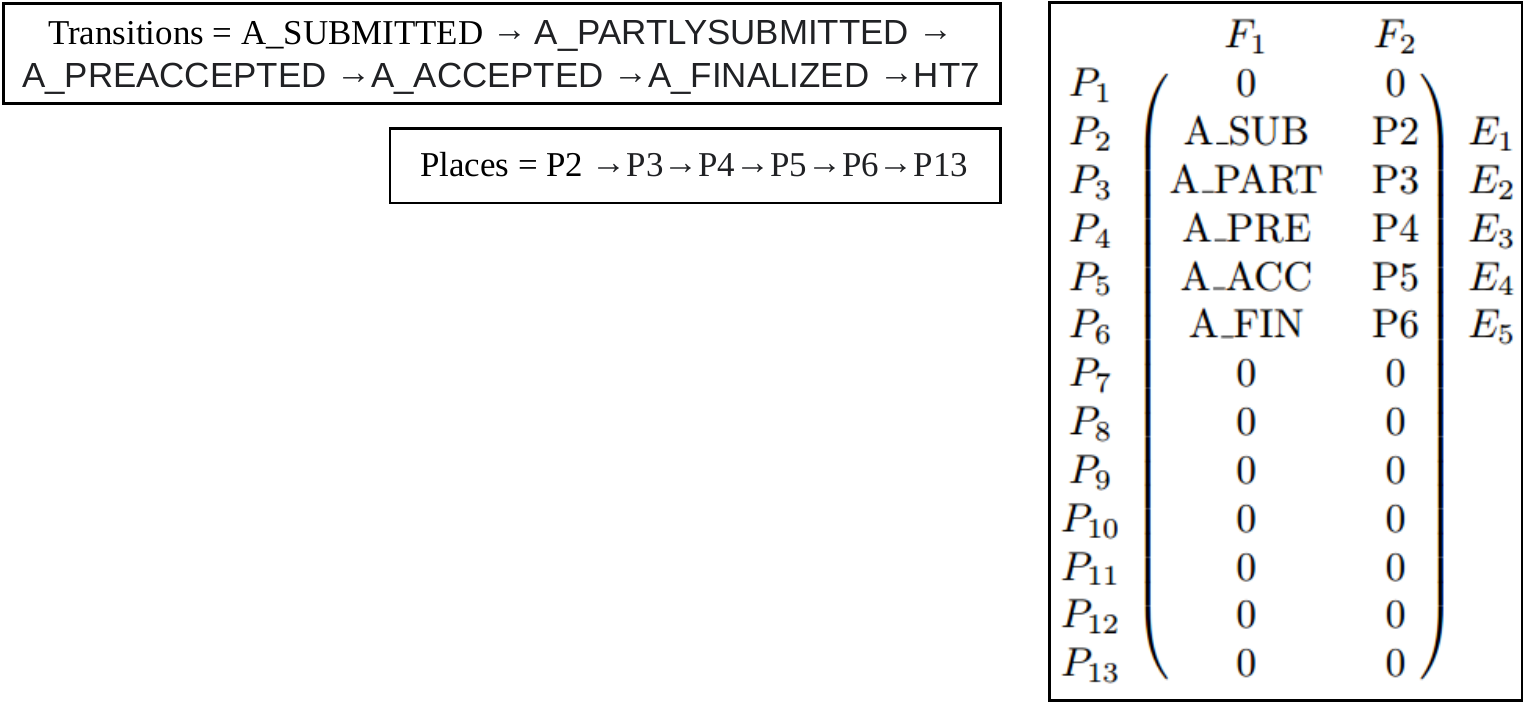}
    \caption{Node feature matrix after encoding the prefix shown on \tablename~\ref{tab:log}. Features are shown as categorical for the sake of clarity.}\label{fig:node-feature-matrix}
\end{figure}

\figurename~\ref{fig:node-feature-matrix} shows an example of the node feature matrix obtained for the replaying of the prefix shown in \tablename~\ref{tab:log} in the process model of \figurename~\ref{fig:model}. TACO is not limited to just using the information from a single node feature matrix, but it takes advantage of the whole sequence of node feature matrices from each event of the whole prefix. Furthermore, TACO also uses the information contained in the events of the log, such as the activity, timestamp or attributes. This information is encoded in a matrix that is called attribute feature matrix. Formally, it can be defined as follows:

\begin{definition}[Attribute feature matrix]
  \label{def:attribute-matrix}
Recalling from Definition \ref{def:event}, let $D_1 \times D_2 \times \dots \times D_m$ be the cross product of the $m$ attribute universes included in a trace, $m > 0$. 
The attribute feature matrix is a matrix $\mathbb{R}^{k \times m}$, 
such that $(a_{ij}) \in D_j \cup \emptyset$,  where $k$ is the length of the longest trace of the event log, $1 \le i \le k$, and $1 \le j \le m$.
\end{definition} 

In particular, this matrix contains the following features: \textit{(i)} the activity whose firing generates a token in the places; \textit{(ii)} the encoded bucket of the time since the previous event; \textit{(iii)} the encoded bucket of the time since the first event of the prefix; and \textit{(iv)} the attributes associated with the event.

\begin{figure}[t]
  \centering
    \includegraphics[width=0.49\textwidth]{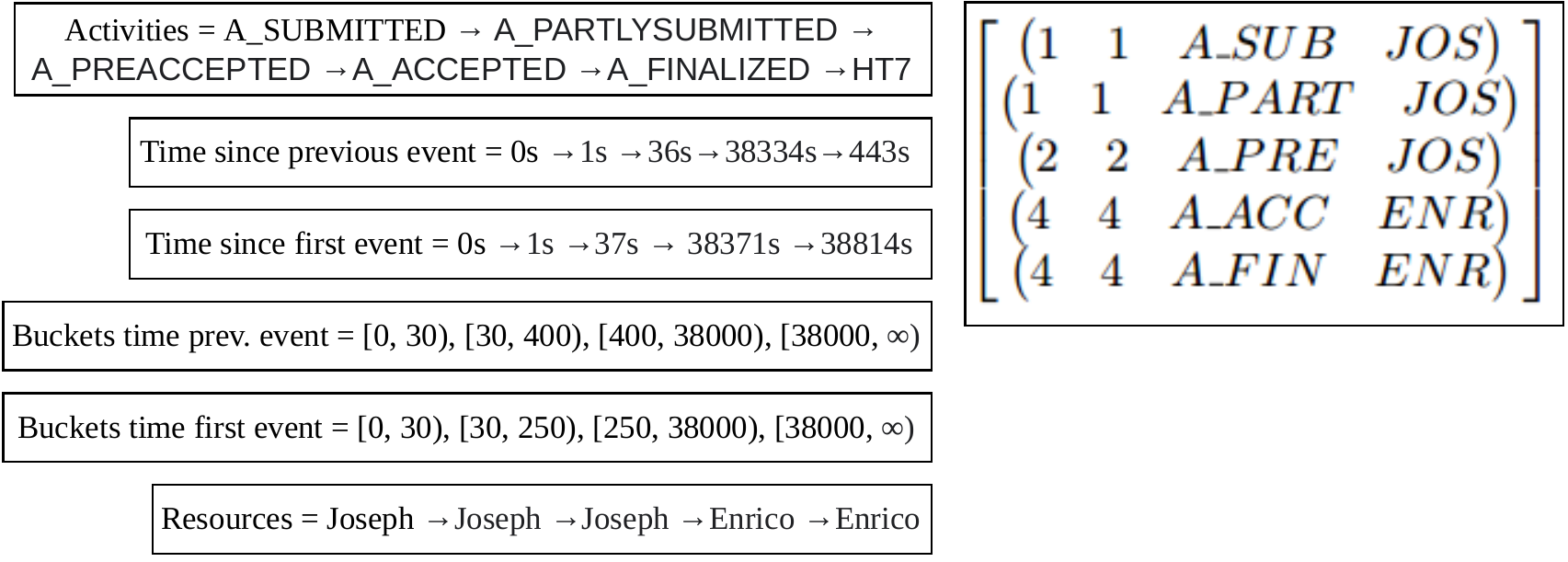}
    \caption{Attribute feature matrix after encoding the prefix from \tablename~\ref{tab:log}. Features are shown as categorical for the sake of clarity.}\label{fig:attribute-feature-matrix}
\end{figure}

\figurename~\ref{fig:attribute-feature-matrix} shows the example of the attribute feature matrix for the prefix of \tablename~\ref{tab:log}. Note that all features in every matrix used by our approach are categorical. To convert continuous features into categorical, we divide the set of continuous values into equal buckets of values and assign each continuous value to one bucket (quantization). Rather than converting the categorical features to one-hot encoding, we separately embed each of the categorical features. This allows to keep the dimensionality of $X$ independent of the number of activities of the event log and to further reduce the memory consumption of the neural network. We use an embedding dimension of 32 for each feature, i.e, the size of the embeddings is of length 32. Also note that the values for each column of the matrix are the accumulative values obtained by the token-replay from the first event of the prefix to the current event. This means that the place information from previous events is not removed.

\subsection{Loop overwriting}
\label{fig:architecture}
\begin{figure}
  \centering
  \includegraphics[width=0.5\textwidth]{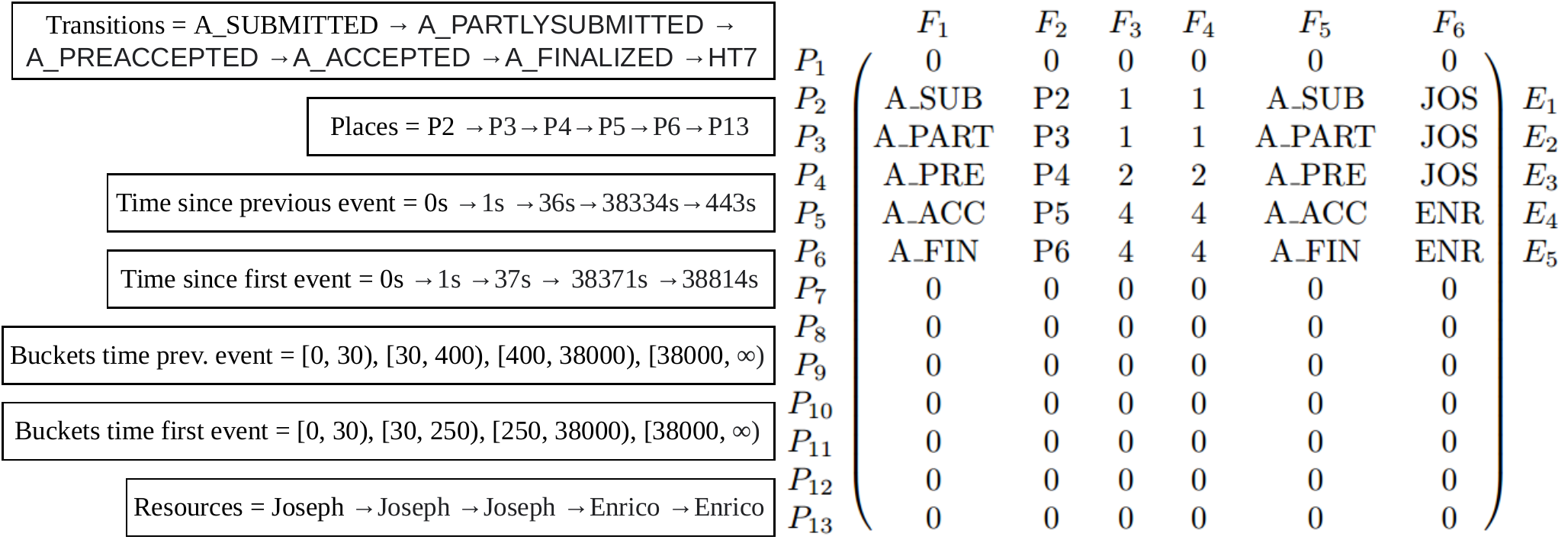}
  \caption{Encoding the prefix from \tablename~\ref{tab:log} by combining the node feature matrix with the attribute feature matrix. Features are shown as categorical for the sake of clarity.}~\label{fig:encoding}
\end{figure}

It could be thought that both the attribute feature matrix and the node feature matrix could be joined in a same matrix, and then apply a GCN that combines this matrix with the normalized random-walk Laplacian. Let \figurename~\ref{fig:encoding} be an example of this approach. In this encoding, the number of equal buckets for the time features is 4 instead of 10 for the sake of simplicity. The element $X_{ij}$ represents the $j$ feature from the place $i$. In this example we assume that the maximum number of transitions fired for each node is only one. Note that the transition HT7 is not fired because we do not know \textit{a priori} whether the prefix has already finished its execution or not.

\begin{figure}
  \centering
  \includegraphics[width=0.45\textwidth]{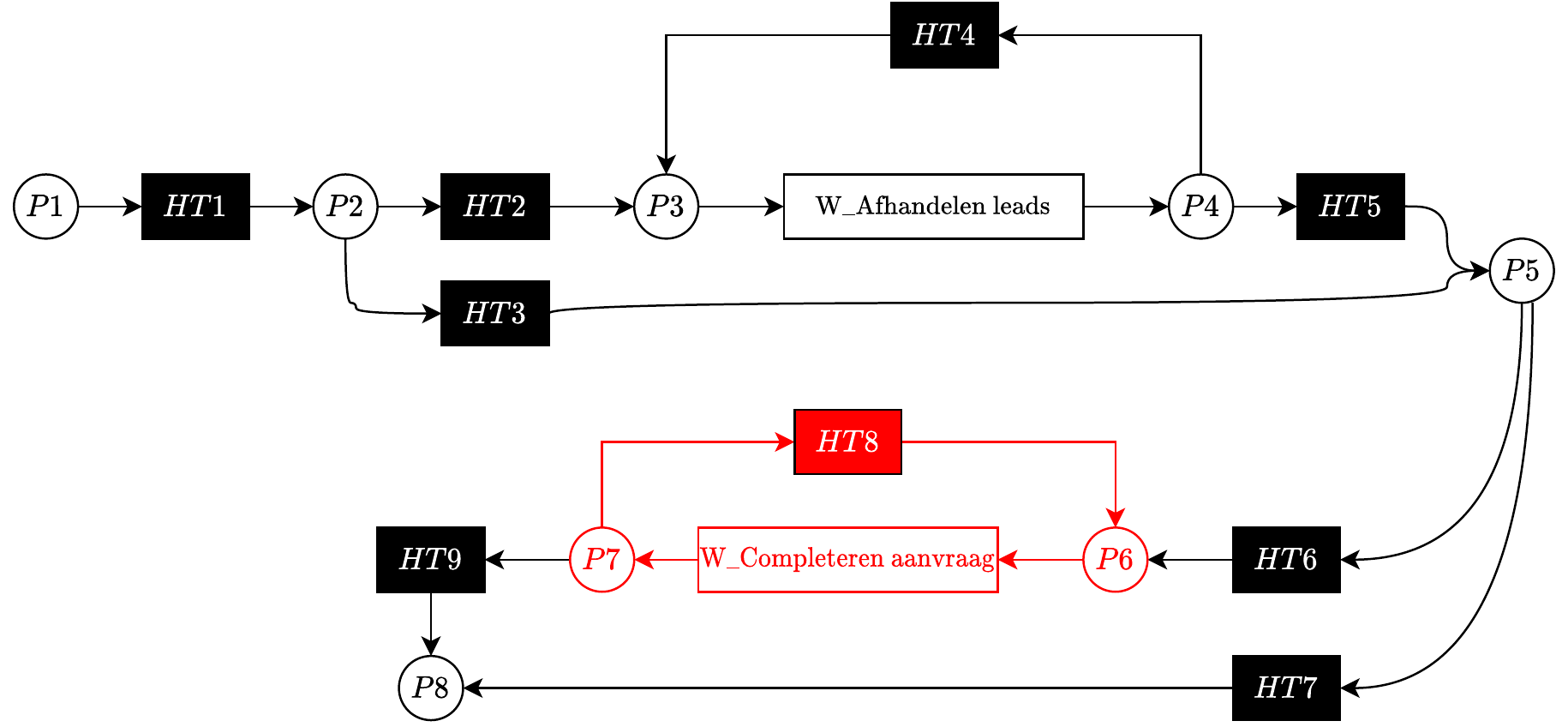}
  \caption{Process model mined from the BPI 2012 W Complete event log.}~\label{fig:overwriting-model}
\end{figure}

\begin{figure}
  \centering
  \includegraphics[width=0.5\textwidth]{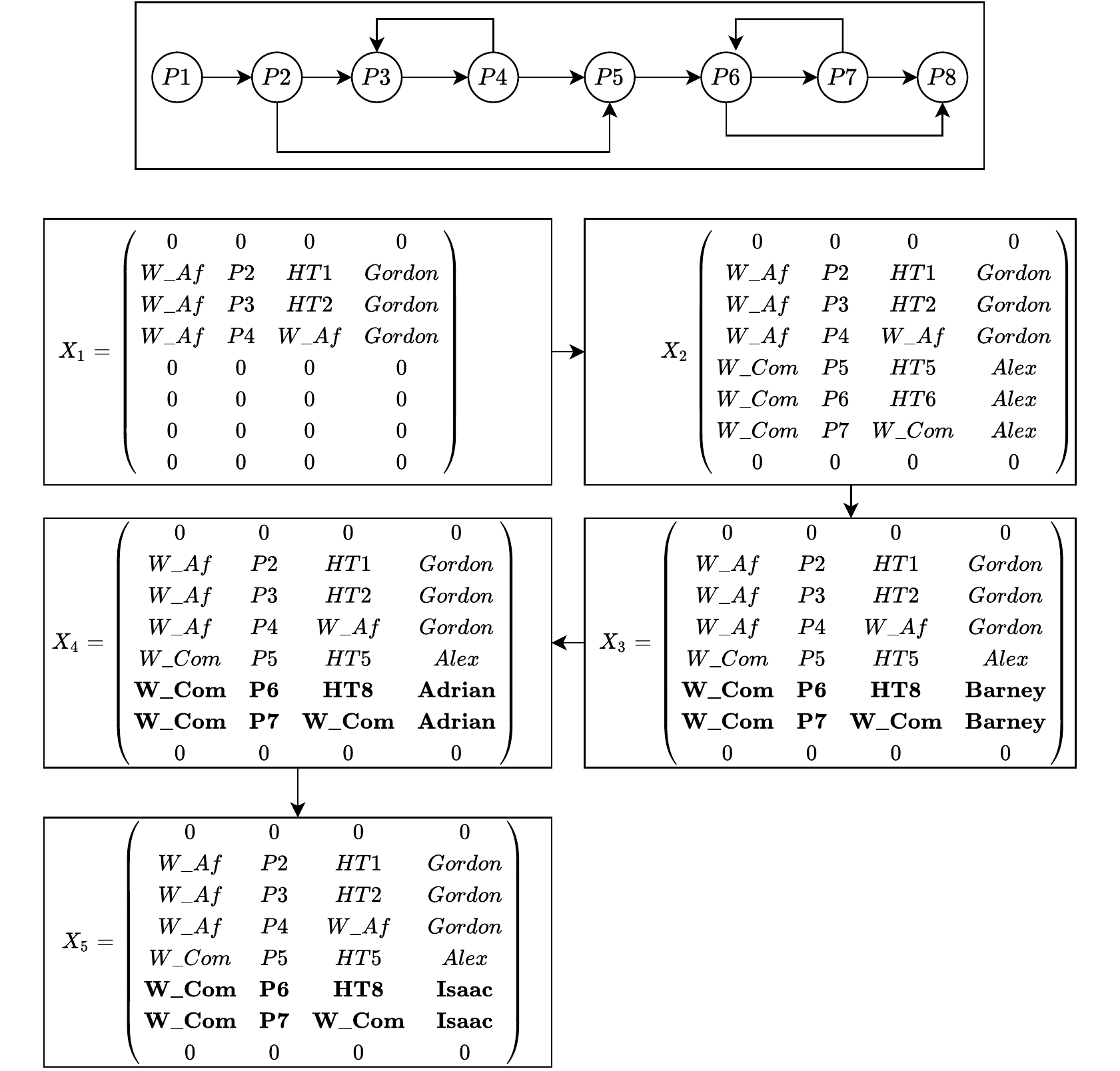}
  \caption{Example of the overwriting problem related to the proposed encoding of the proposal. Time features omitted for the sake of clarity.}~\label{fig:overwriting-example}
\end{figure}

The previous approach has the problem of overwriting the features of nodes when a loop occurs in a set of places. This is specially relevant when the loop consists of only a single activity, because in this case the loss of information is maximum, due to the low number of places involved in this type of loops. Consider the event log BPI Challenge 2012 W Complete~\cite{Dongen2012}, which is a subprocess related to an application of a financial institution~\cite{Dongen2012}. A sample prefix of this event log and its corresponding process model are shown in \tablename~\ref{tab:trace-2012w} and \figurename~\ref{fig:overwriting-model}, respectively. The effect of the loop overwriting is exemplified in \figurename~\ref{fig:overwriting-example}, which shows, on the top, the place graph for the process model depicted in~\figurename~\ref{fig:overwriting-model}, and, on the bottom, the resulting feature matrices obtained from replaying each of the events of \tablename~\ref{tab:trace-2012w}. As shown in this~\figurename, the information of the feature matrix $X_3$ is overwritten in the feature matrix $X_4$, due to the presence of a loop in the activity \textit{W\_Completeren aanvraag}. Assuming a simple GCN model, our only inputs would be the adjacency matrix and the $X_5$ feature matrix.


\begin{table}
  \centering
  \begin{tabular}{|cccc|}
    \hline
    Trace ID & Activity & Timestamp & Resource \\ \hline
    173712 & W\_Afhandelen leads & 01/10/2011 & Gordon \\ \hline
    173712 & W\_Completeren aanvraag & 02/10/2011 & Alex \\ \hline
    173712 & W\_Completeren aanvraag & 03/10/2011 & Barney \\ \hline
    173712 & W\_Completeren aanvraag & 04/10/2011 & Adrian \\ \hline
    173712 & W\_Completeren aanvraag & 05/10/2011 & Isaac \\ \hline
    \ldots & \ldots & \ldots & \ldots \\ \hline
  \end{tabular}
    \caption{Example prefix extracted from the BPI Challenge 2012 W Complete event log.(timestamps have been modified)}\label{tab:trace-2012w}
\end{table}

This encoding problem could be avoided by extending the feature matrix along the feature dimension $F$ up to the maximum loop size present in the event log ($L$) to take into account older events that occurred in the same place, i.e, the feature matrix $X$ would have dimensions $X \in \mathbb{R}^{N \times (F \cdot L_M)}$, where $L_M$ is the maximum loop length. However, some event logs contain loops with a maximum length that would make this approach computationally infeasible.

A naive solution to the problem of loop overwriting would be applying the GCN operation independently to each of the node matrices $X_1, \cdots, X_n$ for each of the events of the prefix. Then, performing the readout operation after the convolution and, finally, feeding that information to a LSTM. However, this solution poses another problem: the structural properties of the input ---the sequence of node feature matrices--- and its temporal properties ---the sequence of attribute feature matrices--- are processed independently and, thus, some spatio-temporal interactions could be missed, such as dependencies within a loop. Therefore, in our approach we propose to use a GRNN, as explained in Section~\ref{sec:grnn}, to simultaneously tackle the spatial and temporal features from the replay of the prefix over the process model. Note that the GRNN only processes the node feature matrix whereas an independent LSTM combines the sequence attribute feature matrix with the embeddings learned by the GRNN.

\section{Evaluation}
\label{sec:evaluation}
\subsection{Experimental setup}

To perform the evaluation of TACO, we relied on the experimentation from~\cite{RamaManeiro2020} under the same exact conditions: a 5-fold cross-validation is performed, splitting the training fold into an 80\%-20\% trace distribution to obtain a validation set. The validation set is used to find the best performing model, in terms of the epoch with the highest validation accuracy. Furthermore, we used the same exact attributes, and we compared the approaches using the ``accuracy'' metric, as reported in~\cite{RamaManeiro2020}. Note that if two events have the same activity label and a different ``lifecycle:transition'' attribute, they are treated as a pair of different activities.

Moreover, we also used the same event logs, but ``Sepsis'' and ``Nasa'', because they do not have resources and, therefore, they can not be applied to the approach~\cite{Camargo2019}. The characteristics of these event logs are depicted in \tablename~\ref{tab:log-info}.  As \tablename~\ref{tab:log-info} shows, most of these event logs have a high variability in both the temporal characteristics of the process execution ---duration of events and traces--- and in the length of the traces. There are two modes of executing the approach of~\cite{Theis2019}, so the two configurations are reported. Furthermore, not only we tested our approach against the original proposals of~\cite{RamaManeiro2020}, but we extended the experimentation by adding the approaches of~\cite{Venugopal2021, Zararah2021} tested under the same conditions. We used the ``weighted'' variant of~\cite{Venugopal2021} due to having the most consistent results across all the datasets. 

\begin{table*}[htbp]
  \scriptsize
  \centering
\begin{tabular}{l|cccScScScScScScSc}
Event log & \rotatebox{90}{Traces} & \rotatebox{90}{Activities} & \rotatebox{90}{Events} & \rotatebox{90}{\shortstack[l]{Avg. case \\ length}} &
\rotatebox{90}{\shortstack[l]{Max. case \\ length}} & \rotatebox{90}{\shortstack[l]{Avg. event \\ duration}} &
\rotatebox{90}{\shortstack[l]{Max. event \\ duration}} & \rotatebox{90}{\shortstack[l]{Avg. case \\ duration}} &
\rotatebox{90}{\shortstack[l]{Max. case \\ duration}} & \rotatebox{90}{Variants} \\ \hline
Helpdesk & 4580 & 14 & 21348 & 4.66 & 15 & 11.16 & 59.92 & 40.86 & 59.99 & 226
\\
BPI 2012 & 13087 & 36 & 262200 & 20.04 & 175 & 0.45 & 102.85 & 8.62 & 137.22 &
4366 \\
BPI 2012 Complete & 13087 & 23 & 164506 & 12.57 & 96 & 0.74 & 30.92 & 8.61 &
91.46 & 4336 \\
BPI 2012 W & 9658 & 19 & 170107 & 17.61 & 156 & 0.7 & 102.85 & 11.69 & 137.22 &
2621 \\
BPI 2012 W Complete & 9658 & 6 & 72413 & 7.5 & 74 & 1.75 & 30.92 & 11.4 & 91.04
& 2263 \\
BPI 2012 O & 5015 & 7 & 31244 & 6.23 & 30 & 3.28 & 69.93 & 17.18 & 89.55 & 168
\\
BPI 2012 A & 13087 & 10 & 60849 & 4.65 & 8 & 2.21 & 89.55 & 8.08 & 91.46 & 17
\\
BPI 2013 closed problems & 1487 & 7 & 6660 & 4.48 & 35 & 51.42 & 2254.84 &
178.88 & 2254.85 & 327 \\
BPI 2013 incidents & 7554 & 13 & 65533 & 8.68 & 123 & 1.57 & 722.25 & 12.08 &
771.35 & 2278 \\
Env. permit & 1434 & 27 & 8577 & 5.98 & 25 & 1.09 & 268.97 & 5.41 &
275.84 & 116 \\ \hline
\end{tabular}
\caption{Statistics of the event logs used for benchmarking. Time related
  measures are shown in days.}~\label{tab:log-info}
\end{table*}

Regarding the mined process models, \tablename~\ref{tab:model-info} shows the average statistics of the mined models for the cross-validation train and validation sets. The reported metrics are the number of loops available in the whole process model ---\textit{loops}---, the number of edges in the process model ---\textit{edges}---, the number of transitions and places of the process model ---\textit{transitions} and \textit{places}, respectively---, the maximum length of the loops that contain exclusively the same activity ---\textit{max. rep. loop}---, and the average length of the loops that contain exclusively the same activity ---\textit{avg. rep. loop}---. This table shows that the mined process models vary on complexity. The most simple process models are the ones mined from the event logs \textit{BPI 2012 A} and \textit{BPI 2012 O}, since they have the lowest number of loops, transitions and places; while the most complex event logs are the \textit{BPI 2012}, \textit{BPI 2012 Complete} and \textit{Env. permit}. Note that the \textit{Complete} variants of the BPI 2012 logs have more loops than their non-complete counterparts since we treat each combination of \textit{lifecycle:transition} and activity as a separate activity.

We configured our approach with the same hyperparameters for every event log. We used 256 hidden units for both the LSTM and the GRNN, and the Adam optimizer with a learning rate of 1e-3 with a cosine annealing warm restart scheduler. We also used 10 buckets to quantize the time features and an embedding dimension of size 32. Furthermore, no hyperparameter tuning was performed since we found that our approach is very resilient to a wide array of different hyperparameters. Instead, we fixed the same hyperparameters for every tested event log.  All experiments have been carried out in an Intel Xeon Gold 5220 equipped with a Tesla V100S. We implemented our approach in PyTorch 1.8.1 relying on Pm4Py~\cite{Berti2019} for processing the event logs.

\begin{table}[htbp]
\scriptsize
\centering
\begin{tabular}{l|cScScScScSc}
  Event log & \rotatebox{90}{Loops} & \rotatebox{90}{Edges} & \rotatebox{90}{Transitions} & \rotatebox{90}{Places} & \rotatebox{90}{Max. rep. loop} & \rotatebox{90}{Avg. rep. loop} \\ \hline
 BPI 2012 & 18404807.6 & 381.2 & 190.6 & 68 & 3 & 2 \\ \hline
 BPI 2012 A & 5 & 73.6 & 36.8 & 19 & 0 & 0 \\ \hline
 BPI 2012 Complete & 18404807.6 & 384.4 & 192.2 & 69.6 & 56 & 4.73 \\ \hline
 BPI 2012 O & 9 & 48 & 24 & 13 & 0 & 0 \\ \hline
 BPI 2012 W & 15.6 & 95.6 & 47.8 & 28.0 & 3 & 2 \\ \hline
 BPI 2012 W Complete & 15.6 & 96 & 48 & 28.2 & 57 & 4.66 \\ \hline
 BPI 2013 closed problems & 50 & 96.4 & 48.2 & 28.2 & 5 & 2.16 \\ \hline
 Helpdesk & 571.6 & 92.8 & 186.4 & 45.6 & 5 & 2.1 \\ \hline
 BPI 2013 incidents & 128.4 & 65.2 & 32.6 & 16 & 19 & 2.07\\ \hline
 Env. permit & 83187.2 & 255.2 & 127.6 & 52.2 & 3 & 3\\ \hline
\end{tabular}
\caption{Average statistics about the mined process models.}\label{tab:model-info}
\end{table}


We performed a two-stage statistical comparison with the aim of reducing the number of statistical tests to make and to ease the interpretation of the results. We decided to use Bayesian statistical tests instead of the classical Null Hypothesis Statistical Tests (NSHT) because they are easier to interpret, and are more powerful in quantifying the differences between the approaches. First, a Bayesian approach to rank models based on the Plackett-Luce model~\cite{Calvo2018} is applied. Then, a Bayesian hierarchical test~\cite{Benavoli2017} between our approach and the other two best approaches according to the previous ranking is performed. Note that the usefulness of the hierarchical Bayesian test resides in that the test uses the full accuracy results from the individual folds of the cross-validation testing technique, so it accounts the fact that one approach can perform badly in one fold but very well on the rest. We rely on the R library \texttt{scmamp} to perform these statistical tests~\cite{Calvo2016}. 

\subsection{Results}

\tablename~\ref{tab:results} shows the results of the evaluation. TACO obtains the best result in 7 of the tested event logs, the second-best result in one of them, and the third-best in two of them. In general, TACO outperforms the other approaches but in event logs that have a simple process model, namely, the \textit{BPI 2012 A} and the \textit{BPI 2012 O}, so a regular RNN model is sufficient to capture the dependencies between the events. These results would confirm that the graph-based approach facilitates the identification of the behavioral patterns, which would help to predict the next activity, mainly when loops and/or parallels are present in the event log. Furthermore, even though TACO underperforms on the \textit{BPI 2012 W Complete}, it still obtains better results than most approaches, apart from the variations from Theis et al.

\definecolor{deepsaffron}{rgb}{1.0, 0.6, 0.2}
\begin{table*}
\centering
\scriptsize
\begin{tabular}{p{2cm}|cccccccccc}

{} &     \rotatebox{90}{BPI 2012} &   \rotatebox{90}{BPI 2012 A} & \rotatebox{90}{\shortstack[l]{BPI 2012 \\ Complete}} &   \rotatebox{90}{BPI 2012 O} &   \rotatebox{90}{BPI 2012 W} & \rotatebox{90}{\shortstack[l]{BPI 2012 \\ W Complete}} & \rotatebox{90}{\shortstack[l]{BPI 2013 \\ Closed Problems}} & \rotatebox{90}{\shortstack[l]{BPI 2013 \\ Incidents}} &             \rotatebox{90}{Env Permit} &               \rotatebox{90}{Helpdesk} \\ \hline

Camargo & 83.28 & 75.98 & 77.93 & 81.35 & 76.4 & 68.95 & 54.67 & 66.68 & 85.78 & 82.93 \\\hline
Evermann & 59.33 & 75.82 & 62.38 & 79.42 & 75.37 & 67.53 & 58.83 & 66.78 & 76.19 & 83.66 \\\hline
Hinkka & \cellcolor{deepsaffron}{\textbf{86.65}} & \cellcolor{cyan}{\textbf{81.19}} & \cellcolor{deepsaffron}{\textbf{80.64}} & \cellcolor{cyan}{\textbf{87.23}} & 84.78 & 70.54 & \cellcolor{yellow}{\textbf{63.47}} & \cellcolor{deepsaffron}{\textbf{74.69}} & 84.43 & 83.08 \\\hline
Khan & 42.9 & 74.9 & 47.37 & 66.08 & 60.15 & 52.22 & 43.58 & 51.91 & 83.59 & 79.97 \\\hline
Mauro & 84.66 & 79.76 & 80.06 & \cellcolor{yellow}{\textbf{82.74}} & 85.98 & 68.64 & 24.94 & 36.67 & 53.59 & 31.79 \\ \hline
Pasquadibisceglie & 83.25 & 74.12 & 74.6 & 78.88 & 81.19 & 68.34 & 47.45 & 46.03 & \cellcolor{deepsaffron}{\textbf{86.69}} & 83.93 \\\hline
Tax & \cellcolor{yellow}{\textbf{85.46}} & 79.53 & \cellcolor{yellow}{\textbf{80.38}} & 82.29 & 85.35 & 69.79 & \cellcolor{deepsaffron}{\textbf{64.01}} & \cellcolor{yellow}{\textbf{70.09}} & 85.71 & \cellcolor{deepsaffron}{\textbf{84.19}}\\ \hline
Theis et al. (w/o attributes) & 82.89 & 65.5 & 75.26 & 78.38 & \cellcolor{yellow}{\textbf{86.22}} & \cellcolor{deepsaffron}{\textbf{80.06}} & 59.48 & 59.41 & \cellcolor{yellow}{\textbf{86.29}} & 78.77 \\\hline
Theis et al. (w/ attributes) & 80.96 & 65.67 & 75.75 & 76.89 & \cellcolor{deepsaffron}{\textbf{86.86}} & \cellcolor{cyan}{\textbf{83.84}} & 54.65 & 51.5 & 85.12 & 79.69 \\\hline
Venugopal & 54.69 & 54.88 & 63.75 & 67.87 & 53.9 & 64.8 & 48.44 & 49.62 & 69.58 & 78.7\\\hline
Zararah & 85.31 & \cellcolor{deepsaffron}{\textbf{79.87}} & 77.43 & 81.15 & 85.19 & 68.19 & 63.29 & 68.94 & 86.06 & \cellcolor{yellow}{\textbf{83.96}} \\ \hline \hline

TACO & \cellcolor{cyan}{\textbf{87.08}} & \cellcolor{yellow}{\textbf{79.78}} & \cellcolor{cyan}{\textbf{80.85}} & \cellcolor{deepsaffron}{\textbf{82.95}} & \cellcolor{cyan}{\textbf{88.34}} & \cellcolor{yellow}{\textbf{74.2}} & \cellcolor{cyan}{\textbf{67.53}} & \cellcolor{cyan}{\textbf{77.72}} & \cellcolor{cyan}{\textbf{87.66}} & \cellcolor{cyan}{\textbf{85.2}} \\ \hline

\end{tabular}
\caption{Mean accuracy of the 5-fold cross-validation. Best, second-best, and third-best approaches are highlighted in cyan, orange and yellow, respectively.}\label{tab:results}
\end{table*}

\begin{figure*}[t]
  \centering
  \includegraphics[width=0.8\textwidth]{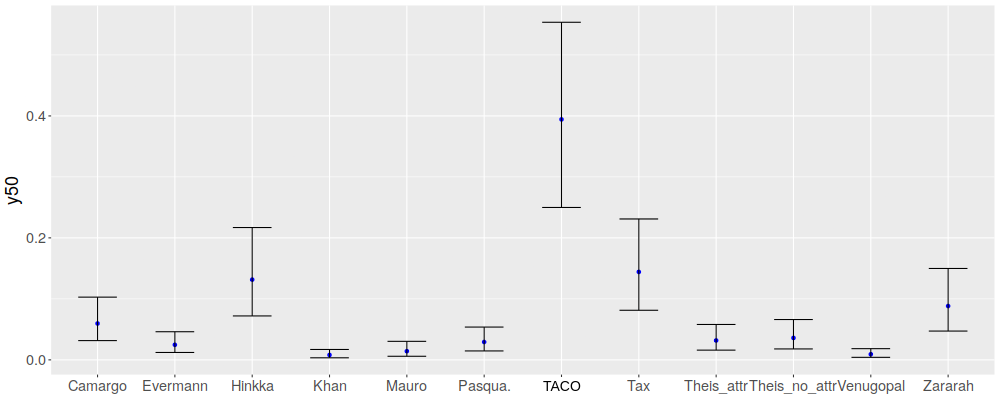}
  \caption{Credible intervals (5\%, 95\% quantiles) and expected probability of winning for the tested approaches.}\label{fig:posteriors}
\end{figure*}

\tablename~\ref{tab:ranking} shows the results of applying the Bayesian Plackett-Luce model for ranking the algorithms. Our approach obtains the best rank overall and the highest probability of being the best approach, with a difference of 25 percentage points over the second one. \figurename~\ref{fig:posteriors} shows the credible intervals --- 5\% and 95\% quantiles --- as well as the expected probability of winning for every tested approach. Note that a non-overlapping pair of approaches means that they are statistically different. The credible interval of the GRNN approach does not overlap any other one, which shows that our approach is statistically different from the non-overlapping ones.

\begin{table}[t]
  \centering
  \scriptsize
\begin{tabular}{|c|cQcQcQcQcQcQcQcQcQcQcQc|}
\hline
     & \rotatebox{90}{TACO}  & \rotatebox{90}{Tax} & \rotatebox{90}{Hinkka} & \rotatebox{90}{Zararah} & \rotatebox{90}{Camargo} & \rotatebox{90}{Theis (w/o)} & \rotatebox{90}{Theis (w/)} & \rotatebox{90}{Pasqua.} & \rotatebox{90}{Evermann} & \rotatebox{90}{Mauro} & \rotatebox{90}{Venugopal} & \rotatebox{90}{Khan} \\ \hline
Rank & 1.03 & 2.59 & 2.83 & 3.92 & 5.07 & 6.81 & 7.31 & 7.62 & 8.27 & 10.02 & 11.14 & 11.4 \\ \hline
Prob. (\%) & 39.7 & 14.9 & 13.6 & 9.2 & 6.3 & 3.8 & 3.4 & 3.1 & 2.6 & 1.6 & 1 & 0.9 \\ \hline
\end{tabular}
\caption{Plackett-Luce rankings and posterior probabilities of the approaches.}~\label{tab:ranking}
\end{table}

To further assess the differences of the approaches, we apply a hierarchical Bayesian test whose results are shown in the \tablename~\ref{tab:hierarchical-tax} and \tablename~\ref{tab:hierarchical-hinkka}. These tables, given two approaches $A$ and $B$, show the probability of $A$ being better than $B$ ($A > B$), the probability of the two approaches being equal ($A = B$), and the probability of $A$ being worse than $B$ ($A > B$). Furthermore, we can assess whether $A$ is not worse than $B$ by summing the probabilities from $A > B$ and $A = B$. In this test, it is assumed that a probability greater than 95\% means statistically significance. We perform this test to compare our approach against the two best ones according to the Plackett-Luce ranking: Hinkka et al.~\cite{Hinkka2019} and Tax et al.~\cite{Tax2017}. These tables report both the probabilities for each individual dataset and the overall probability of being better.

On the one hand, \tablename~\ref{tab:hierarchical-tax} compares TACO against the one of Tax et al., which means that our approach significantly outperforms Tax et al. overall. Furthermore, we significantly outperform Tax et al. in every event log except in the event logs \textit{BPI 2012 A} and \textit{BPI 2012 O}, which are the simpler ones in terms of process model according to the statistics from \tablename~\ref{tab:model-info}. This highlights the fact that a simple RNN is enough to capture every dependence between the events when the underlying process model are simple enough.

\begin{table}[htbp]
    \centering
    \scriptsize
    \begin{tabular}{l|ccc}
      Event log & $A > T$ & $A = T$ & $A < T$ \\ \hline
      BPI 2012 & 99.91\% & 0.08\% & 0.007\% \\ 
      BPI 2012 A & 75.32\% & 23.33\% & 1.35\% \\ 
      BPI 2012 Complete & 98.08\% & 1.61\% & 0.31\% \\ 
      BPI 2012 O & 84.12\% & 15.20\% & 0.68\% \\ 
      BPI 2012 W & 99.99\% & 0.002\% & 0\% \\ 
      BPI 2012 W Complete & 99.90\% & 0.09\% & 0.004\% \\ 
      BPI 2013 closed problems & 99.96\% & 0.036\% & 0.001\% \\ 
      BPI 2013 incidents & 99.95\% & 0.046\% & 0.001\% \\ 
      Env. permit & 99.06\% & 0.88\% & 0.05\% \\
      Helpdesk & 99.54\% & 0.42\% & 0.043\% \\ \hline \hline
      \textbf{Overall} & 99.4\% & 0\% & 0.6\% \\ \hline
    \end{tabular}
    \caption{Hierarchical Bayesian tests per dataset: TACO (A) vs Tax et al. (T)}\label{tab:hierarchical-tax}
\end{table}

On the other hand, \tablename~\ref{tab:hierarchical-hinkka} compares TACO with Hinkka et al. In this case, it outperforms Hinkka et al. with a 92\% probability, falling less than 3\% short of statistical significance. In fact, it outperforms siginificantly Hinkka et al. in six event logs: \textit{BPI 2012}, \textit{BPI 2012 W Complete}, \textit{BPI 2012 W}, \textit{BPI 2013 incidents}, \textit{Env. permit}, and \textit{Helpdesk}. The first four event logs are among the most complex ones, as \tablename~\ref{tab:log-info}, and the last two, even though they are shorter, they still have structural complexity. Furthermore, TACO is not worse than Hinkka et al. in the event logs \textit{BPI 2012 Complete}, and \textit{BPI 2013 closed problems} ($G > H$ greater than 90\% of the cases). Finally, TACO is equal to Hinkka et al. in the \textit{BPI 2012 A}, and \textit{BPI 2012 O}, which are the simpler event logs used. Note that the approach of Hinkka et al. does not outperform us in any event log nor is worse than GRNN.

\begin{table}[htbp]
  \centering
  \scriptsize
    \begin{tabular}{l|ccc}
      Event log & $A > H$ & $A = H$ & $A < H$ \\ \hline
      BPI 2012 & 99.37\% & 0.52\% & 0.1\% \\
      BPI 2012 A & 0.09\% & 99.90\% & 0.005\% \\
      BPI 2012 Complete & 94.84\% & 3.76\% & 1.40\% \\
      BPI 2012 O & 0.15\% & 99.85\% & 0.002\% \\
      BPI 2012 W & 100\% & 0\% & 0\% \\ 
      BPI 2012 W Complete & 99.96\% & 0.036\% & 0.0008\% \\
      BPI 2013 c.p. & 90.61\% & 9.17\% & 0.214\% \\
      BPI 2013 incidents & 99.83\% & 0.164 & 0.008\% \\
      Env. permit & 99.74\% & 0.251\% & 0.006\% \\
      Helpdesk & 99.95\% & 0.041\% & 0.003\% \\ \hline \hline
      \textbf{Overall} & 92\% & 0\% & 8\% \\ \hline
    \end{tabular}
    \caption{Hierarchical Bayesian tests per dataset: TACO (A) vs. Hinkka et al. (H)}\label{tab:hierarchical-hinkka}
\end{table}

\section{Conclusions and future work}
\label{sec:conclusions-future-work}
In this paper we presented TACO, an approach based on GCNs and RNNs that, contrary to most approaches, solves the next activity prediction problem by using, not only the information readily available in the traces of the event log, but also the information available in the process models. For that, we proposed a novel encoding scheme that leverages both the structural information from the process model and the information available in the events of the event log, while being memory efficient and easy to compute. Moreover, we have presented why the usual encoding used for GCNs falls short for predictive monitoring problems. Finally, we showed how to successfully combine a recurrent model with a graph neural network to take advantage of the aforementioned information.

We evaluated our proposal in 10 real life event logs and compared it against 10 approaches of the state-of-the-art. The results show that TACO works more consistently and obtains better results overall than the other approaches.  This shows the adequacy of using GNNs for predictive process monitoring since their internal structure can better leverage the information of the process model. This is also clearly verified in the results, as TACO outperforms the other approaches in every event log except in those that have a simpler process model.

For future work, we intend to test more different graph convolution operators, to extend the approach to other process models, such as resource models, and to tackle more predictive monitoring tasks such as the remaining time prediction.

\section{Acknowledgments}
This work has received financial support from the Consellería de Educación, Universidade e Formación Profesional (accreditation 2019-2022 ED431G-2019/04), the European Regional Development Fund (ERDF), which acknowledges the CiTIUS - Centro Singular de Investigación en Tecnoloxías Intelixentes da Universidade de Santiago de Compostela as a Research Center of the Galician University System, and the Spanish Ministry of Science and Innovation (grants PDC2021-121072-C21 and PID2020-112623GB-I00). E. Rama-Maneiro is supported by the Spanish Ministry of Education, under the FPU national plan (FPU18/05687).

\section{Biographies}

\vspace*{-2\baselineskip}

\begin{IEEEbiography}[{\includegraphics[width=1in,height=1.25in,clip,keepaspectratio]{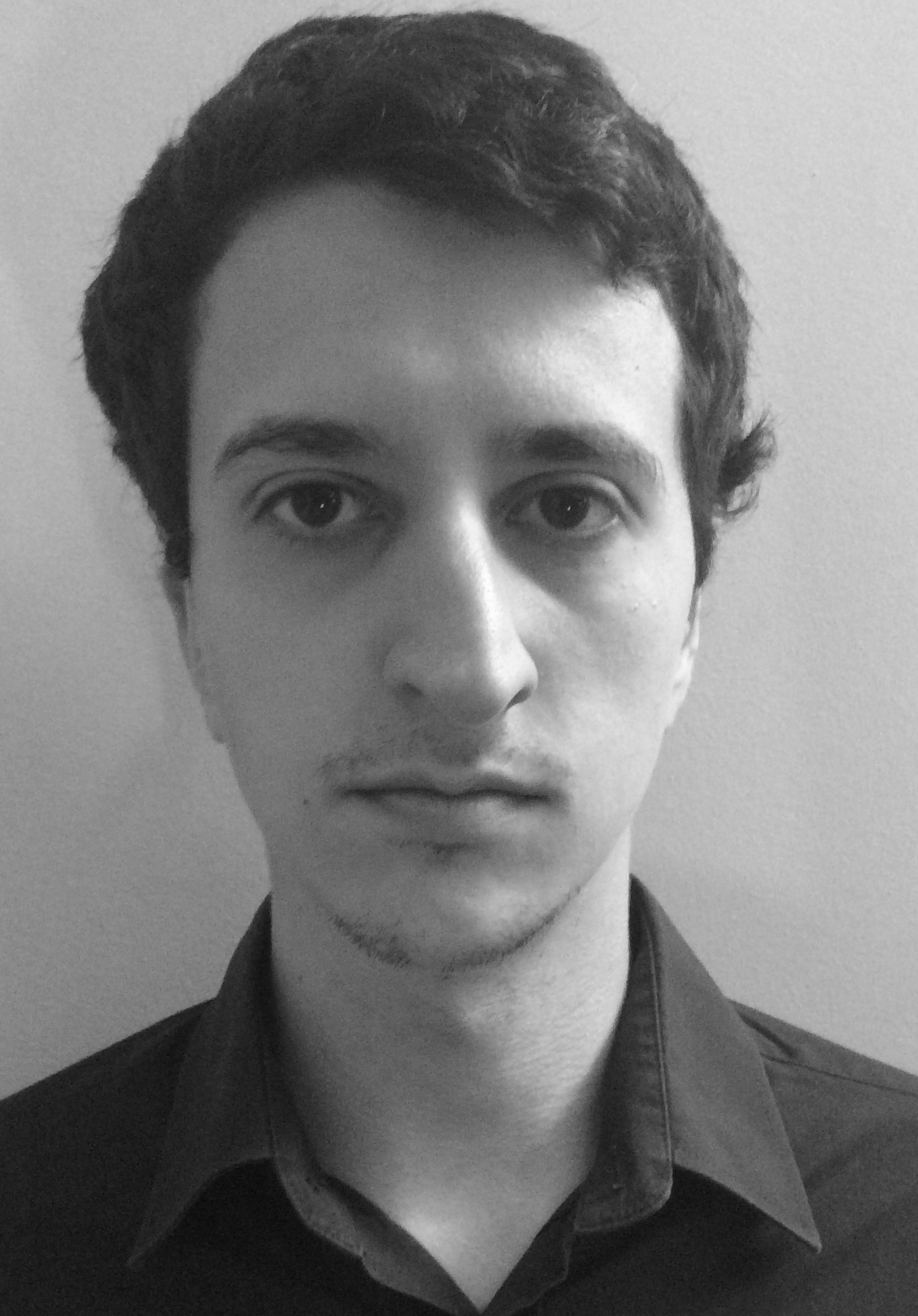}}]{EFRÉN RAMA-MANEIRO}
  received the B.Eng. degree in computer engineering and the M.Sc. degree in Big Data from the University of Santiago de Compostela, Spain in 2018 and 2019 respectively. He is a Researcher and currently working toward the Ph.D. degree at the Centro Singular de Investigación en Tecnoloxías Intelixentes, University of Santiago de Compostela. His research interests include process mining and deep learning.
\end{IEEEbiography}

\vspace*{-2\baselineskip}

\begin{IEEEbiography}[{\includegraphics[width=1in,height=1.25in,clip,keepaspectratio]{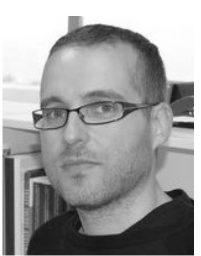}}]{JUAN C. VIDAL}
received the B.Eng. degree in computer science from the University of La Coruña, La Coruña, Spain, in 2000, and the Ph.D. degree in artificial intelligence from the University of Santiago de Compostela (USC), Santiago de Compostela, Spain, in 2010, where he was an Assistant Professor with the Department of Electronics and Computer Science, from 2010 to 2017. He is currently an Associate Researcher with the Centro Singular de Investigación en Tecnoloxías Intelixentes (CiTIUS), USC. His research interests include process mining, fuzzy logic, machine learning, and linguistic summarization.
\end{IEEEbiography}

\vspace*{-2\baselineskip}

\begin{IEEEbiography}[{\includegraphics[width=1in,height=1.25in,clip,keepaspectratio]{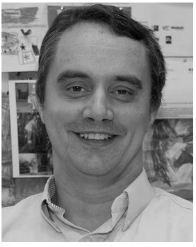}}]{MANUEL LAMA}
received the Ph.D. degree in physics from the University of Santiago de Compostela, in 2000, where he is currently an Associate Professor of artificial intelligence. He has collaborated on more than 30 projects and research contracts financed by public calls, participating as a principal investigator in 20 of them. These activities were implemented in areas, such as process discovery, predictive monitoring, and management of dynamic processes. As a result of this research, he has published over 150 scientific articles with review process in conference and national and international journals.
\end{IEEEbiography}

\bibliographystyle{IEEEtran}
\bibliography{paper}

\begin{thebibliography}{10}
\providecommand{\url}[1]{#1}
\csname url@samestyle\endcsname
\providecommand{\newblock}{\relax}
\providecommand{\bibinfo}[2]{#2}
\providecommand{\BIBentrySTDinterwordspacing}{\spaceskip=0pt\relax}
\providecommand{\BIBentryALTinterwordstretchfactor}{4}
\providecommand{\BIBentryALTinterwordspacing}{\spaceskip=\fontdimen2\font plus
\BIBentryALTinterwordstretchfactor\fontdimen3\font minus
  \fontdimen4\font\relax}
\providecommand{\BIBforeignlanguage}[2]{{%
\expandafter\ifx\csname l@#1\endcsname\relax
\typeout{** WARNING: IEEEtran.bst: No hyphenation pattern has been}%
\typeout{** loaded for the language `#1'. Using the pattern for}%
\typeout{** the default language instead.}%
\else
\language=\csname l@#1\endcsname
\fi
#2}}
\providecommand{\BIBdecl}{\relax}
\BIBdecl

\bibitem{Aalst2012}
W.~M.~P. van der Aalst~et al., ``Process mining manifesto,'' in
  \emph{Proceedings of the 9th International Business Process Management
  Workshops (BPM 2011)}, ser. Lecture Notes in Business Information Processing,
  vol.~99.\hskip 1em plus 0.5em minus 0.4em\relax Springer, 2011, pp. 169--194.

\bibitem{Kirchmer2017}
M.~Kirchmer, \emph{High Performance Through Business Process Management}.\hskip
  1em plus 0.5em minus 0.4em\relax Springer International Publishing, 2017.

\bibitem{Dongen2012}
B.~van Dongen, ``Bpi challenge 2012,'' 2012.

\bibitem{Maggi2014}
F.~M. Maggi, C.~D. Francescomarino, M.~Dumas, and C.~Ghidini, ``Predictive
  monitoring of business processes,'' in \emph{Advanced Information Systems
  Engineering}.\hskip 1em plus 0.5em minus 0.4em\relax Springer International
  Publishing, 2014, pp. 457--472.

\bibitem{Tax2018}
N.~Tax, I.~Teinemaa, and S.~J. van Zelst, ``An interdisciplinary comparison of
  sequence modeling methods for next-element prediction,'' \emph{Software and
  Systems Modeling}, 2020.

\bibitem{RamaManeiro2020}
E.~Rama-Maneiro, J.~C. Vidal, and M.~Lama, ``Deep learning for predictive
  business process monitoring: Review and benchmark,'' Sep. 2020.

\bibitem{Mehdiyev2017}
N.~Mehdiyev, J.~Evermann, and P.~Fettke, ``A multi-stage deep learning approach
  for business process event prediction,'' in \emph{Proceedings of the 2017
  {IEEE} 19th Conference on Business Informatics ({CBI} 2017)}.\hskip 1em plus
  0.5em minus 0.4em\relax {IEEE}, 2017, pp. 119--128.

\bibitem{Mehdiyev2018}
------, ``A novel business process prediction model using a deep learning
  method,'' \emph{Business {\&} Information Systems Engineering}, 2018.

\bibitem{Taymouri2020}
F.~Taymouri and et~al., ``Predictive business process monitoring via generative
  adversarial nets: The case of next event prediction,'' in \emph{Lecture Notes
  in Computer Science}.\hskip 1em plus 0.5em minus 0.4em\relax Springer
  International Publishing, 2020, pp. 237--256.

\bibitem{Mauro2019}
N.~D. Mauro, A.~Appice, and T.~M.~A. Basile, ``Activity prediction of business
  process instances with inception {CNN} models,'' in \emph{Lecture Notes in
  Computer Science}.\hskip 1em plus 0.5em minus 0.4em\relax Springer
  International Publishing, 2019, pp. 348--361.

\bibitem{Pasquadibisceglie2019}
V.~Pasquadibisceglie, A.~Appice, G.~Castellano, and D.~Malerba, ``Using
  convolutional neural networks for predictive process analytics,'' in
  \emph{2019 International Conference on Process Mining ({ICPM})}.\hskip 1em
  plus 0.5em minus 0.4em\relax {IEEE}, jun 2019.

\bibitem{Hinkka2019}
M.~Hinkka, T.~Lehto, and K.~Heljanko, ``Exploiting event log event attributes
  in {RNN} based prediction,'' in \emph{Communications in Computer and
  Information Science}.\hskip 1em plus 0.5em minus 0.4em\relax Springer
  International Publishing, 2019, pp. 405--416.

\bibitem{Khan2018}
A.~Khan, H.~Le, K.~Do, T.~Tran, A.~Ghose, H.~Dam, and R.~Sindhgatta,
  ``Memory-augmented neural networks for predictive process analytics,'' Feb.
  2018.

\bibitem{Heinrich2021}
K.~Heinrich, P.~Zschech, C.~Janiesch, and M.~Bonin, ``Process data properties
  matter: Introducing gated convolutional neural networks ({GCNN}) and
  key-value-predict attention networks ({KVP}) for next event prediction with
  deep learning,'' vol. 143, p. 113494, apr 2021.

\bibitem{Evermann2017}
J.~Evermann, J.-R. Rehse, and P.~Fettke, ``Predicting process behaviour using
  deep learning,'' \emph{Decision Support Systems}, vol. 100, pp. 129--140, aug
  2017.

\bibitem{Tax2017}
N.~Tax, I.~Verenich, M.~L. Rosa, and M.~Dumas, ``Predictive business process
  monitoring with {LSTM} neural networks,'' in \emph{Advanced Information
  Systems Engineering}.\hskip 1em plus 0.5em minus 0.4em\relax Springer
  International Publishing, 2017, pp. 477--492.

\bibitem{Camargo2019}
M.~Camargo, M.~Dumas, and O.~Gonz{\'{a}}lez-Rojas, ``Learning accurate {LSTM}
  models of business processes,'' in \emph{Lecture Notes in Computer
  Science}.\hskip 1em plus 0.5em minus 0.4em\relax Springer International
  Publishing, 2019, pp. 286--302.

\bibitem{Zararah2021}
Z.~A. Bukhsh, A.~Saeed, and R.~M. Dijkman, ``Processtransformer: Predictive
  business process monitoring with transformer network,'' 2021.

\bibitem{Nguyen2019}
H.~T.~C. Nguyen, S.~Lee, J.~Kim, J.~Ko, and M.~Comuzzi, ``Autoencoders for
  improving quality of process event logs,'' \emph{Expert Systems with
  Applications}, vol. 131, pp. 132--147, oct 2019.

\bibitem{Theis2019}
J.~Theis and H.~Darabi, ``Decay replay mining to predict next process events,''
  \emph{{IEEE} Access}, vol.~7, pp. 119\,787--119\,803, 2019.

\bibitem{Venugopal2021}
I.~Venugopal, J.~Tollich, M.~Fairbank, and A.~Scherp, ``A comparison of
  deep-learning methods for analysing and predicting business
  processes.''\hskip 1em plus 0.5em minus 0.4em\relax {IEEE}, jul 2021.

\bibitem{Weinzierl2021}
S.~Weinzierl, ``Exploring gated graph sequence neural networks for predicting
  next process activities,'' in \emph{5th International Workshop on Artificial
  Intelligence for Business Process Management (AI4BPM2021)}, 07 2021.

\bibitem{Graves2016}
A.~Graves and et~al., ``Hybrid computing using a neural network with dynamic
  external memory,'' \emph{Nature}, vol. 538, no. 7626, pp. 471--476, oct 2016.

\bibitem{Jalayer2020}
A.~Jalayer, M.~Kahani, A.~Beheshti, A.~Pourmasoumi, and H.~R. Motahari-Nezhad,
  ``Attention mechanism in predictive business process monitoring.''\hskip 1em
  plus 0.5em minus 0.4em\relax {IEEE}, oct 2020.

\bibitem{Szegedy2015}
C.~Szegedy, W.~Liu, Y.~Jia, P.~Sermanet, S.~Reed, D.~Anguelov, D.~Erhan,
  V.~Vanhoucke, and A.~Rabinovich, ``Going deeper with convolutions,'' in
  \emph{2015 {IEEE} Conference on Computer Vision and Pattern Recognition
  ({CVPR})}.\hskip 1em plus 0.5em minus 0.4em\relax {IEEE}, jun 2015.

\bibitem{Dauphin2017}
Y.~N. Dauphin, A.~Fan, M.~Auli, and D.~Grangier, ``Language modeling with gated
  convolutional networks,'' in \emph{Proceedings of the 34th International
  Conference on Machine Learning, {ICML} 2017, Sydney, NSW, Australia, 6-11
  August 2017}, ser. Proceedings of Machine Learning Research, D.~Precup and
  Y.~W. Teh, Eds., vol.~70.\hskip 1em plus 0.5em minus 0.4em\relax {PMLR},
  2017, pp. 933--941.

\bibitem{Daniluk2017}
M.~Daniluk, T.~Rockt{\"{a}}schel, J.~Welbl, and S.~Riedel, ``Frustratingly
  short attention spans in neural language modeling,'' in \emph{5th
  International Conference on Learning Representations, {ICLR} 2017, Toulon,
  France, April 24-26, 2017, Conference Track Proceedings}, 2017.

\bibitem{Goodfellow2014}
I.~J. Goodfellow and et~al., ``Generative adversarial nets,'' in \emph{Advances
  in Neural Information Processing Systems 27: Annual Conference on Neural
  Information Processing Systems 2014, December 8-13 2014, Montreal, Quebec,
  Canada}, 2014, pp. 2672--2680.

\bibitem{Vaswani2017}
A.~Vaswani, N.~Shazeer, N.~Parmar, J.~Uszkoreit, L.~Jones, A.~N. Gomez,
  L.~Kaiser, and I.~Polosukhin, ``Attention is all you need,'' in
  \emph{Proceedings of the 30th Annual Conference on Neural Information
  Processing Systems ({NIPS} 2017)}, 2017, pp. 5998--6008.

\bibitem{Desel1998}
J.~Desel and W.~Reisig, ``Place/transition petri nets,'' in \emph{Lectures on
  Petri Nets I: Basic Models}.\hskip 1em plus 0.5em minus 0.4em\relax Springer
  Berlin Heidelberg, 1998, pp. 122--173.

\bibitem{Hochreiter1997}
S.~Hochreiter and J.~Schmidhuber, ``Long short-term memory,'' vol.~9, no.~8,
  pp. 1735--1780, nov 1997.

\bibitem{Cho2014}
K.~Cho and et~al., ``Learning phrase representations using {RNN}
  encoder{\textendash}decoder for statistical machine translation,'' in
  \emph{Proceedings of the 2014 Conference on Empirical Methods in Natural
  Language Processing ({EMNLP})}.\hskip 1em plus 0.5em minus 0.4em\relax
  Association for Computational Linguistics, 2014.

\bibitem{Kipf2017}
T.~N. Kipf and M.~Welling, ``Semi-supervised classification with graph
  convolutional networks,'' in \emph{5th International Conference on Learning
  Representations, {ICLR} 2017}, 2017.

\bibitem{Xu2019}
K.~Xu, W.~Hu, J.~Leskovec, and S.~Jegelka, ``How powerful are graph neural
  networks?'' in \emph{7th International Conference on Learning
  Representations, {ICLR} 2019, New Orleans, LA, USA, May 6-9, 2019}, 2019.

\bibitem{Ruiz2020}
L.~Ruiz, F.~Gama, and A.~Ribeiro, ``Gated graph recurrent neural networks,''
  \emph{{IEEE} Transactions on Signal Processing}, vol.~68, pp. 6303--6318,
  2020.

\bibitem{Augusto2018}
A.~Augusto, R.~Conforti, M.~Dumas, M.~L. Rosa, and A.~Polyvyanyy, ``Split
  miner: automated discovery of accurate and simple business process models
  from event logs,'' \emph{Knowledge and Information Systems}, vol.~59, no.~2,
  pp. 251--284, may 2018.

\bibitem{Berti2019}
A.~Berti, S.~J. van Zelst, and W.~M.~P. van~der Aalst, ``Process mining for
  python (pm4py): Bridging the gap between process- and data science,'' 2019.

\bibitem{Calvo2018}
B.~Calvo, J.~Ceberio, and J.~A. Lozano, ``Bayesian inference for algorithm
  ranking analysis,'' in \emph{Proceedings of the Genetic and Evolutionary
  Computation Conference Companion}.\hskip 1em plus 0.5em minus 0.4em\relax
  {ACM}, jul 2018.

\bibitem{Benavoli2017}
A.~Benavoli, G.~Corani, J.~Demsar, and M.~Zaffalon, ``Time for a change: a
  tutorial for comparing multiple classifiers through bayesian analysis,''
  \emph{J. Mach. Learn. Res.}, vol.~18, pp. 77:1--77:36, 2017.

\bibitem{Calvo2016}
B.~Calvo and G.~Santafé, ``{scmamp: Statistical Comparison of Multiple
  Algorithms in Multiple Problems},'' \emph{{The R Journal}}, vol.~8, no.~1,
  pp. 248--256, 2016.

\end{thebibliography}

\end{document}